\documentclass[11pt]{article}

\usepackage{geometry}  
\usepackage[T1]{fontenc}  
\usepackage[utf8]{inputenc}  
\usepackage{lineno}  
\usepackage{enumitem}  
\usepackage{titling}  
\usepackage{etoolbox}
\usepackage{mathtools}  
\usepackage{titlesec}  
\usepackage{gensymb}   
\usepackage{multirow}
\usepackage{placeins}
\usepackage{bm}
\usepackage{caption}
\usepackage[numbers]{natbib}
\usepackage{pdflscape}

\usepackage[english]{babel}  
\usepackage{amsmath}  
\usepackage{amsfonts}  
\usepackage{amssymb}  
\usepackage{wasysym}  
\usepackage{bbm}  
\usepackage{array}  
\usepackage{verbatim} 
\usepackage{float} 
\usepackage{subcaption}
\usepackage{makecell}  

\usepackage{tabularx}
\usepackage{booktabs}
\usepackage[breaklinks=true,hidelinks]{hyperref}
\usepackage{breakcites}
\usepackage{longtable} 
\usepackage{cleveref}
\usepackage{soul}
\usepackage{xcolor}
\usepackage{dcolumn}

\titleformat{name=\section}{\normalfont\Large\bfseries}{}{0pt}{}
\titleformat{name=\subsection}{\normalfont\large\bfseries}{}{0pt}{}
\titleformat{name=\subsubsection}{\normalfont\normalsize\bfseries}{}{0pt}{}

\newcommand{\optincludegraphics}[2][]{}

\newcommand{\optinput}[1]{}

\newtagform{brackets}{[}{]}
\usetagform{brackets}

\title{Envisioning global urban development with satellite imagery and generative AI}

\begin{document}


\begin{titlepage}

{\noindent\LARGE\bf\thetitle}
\bigskip

\begin{flushleft}\large
    Kailai Sun \textsuperscript{1,{\dag}},
    Yuebing Liang \textsuperscript{2,{\dag}},
    Mingyi He \textsuperscript{3},
    Yunhan Zheng \textsuperscript{4},
    Alok Prakash \textsuperscript{1},
    Shenhao Wang \textsuperscript{1,5,{*}},
    Jinhua Zhao \textsuperscript{3,{*}},
    Alex ``Sandy'' Pentland \textsuperscript{6}
\end{flushleft}

\begin{enumerate}[label=\textbf{\arabic*}]
\item Singapore–MIT Alliance for Research and Technology Centre (SMART), Singapore
\item Department of Urban Planning, Tsinghua University, Beijing, China
\item Department of Urban Studies and Planning, Massachusetts Institute of Technology, Cambridge, MA, USA
\item College of Urban and Environmental Sciences, Peking University, Beijing, China
\item Department of Urban and Regional Planning, University of Florida, Gainesville, FL, USA
\item Stanford Institute for Human-Centered Artificial Intelligence, Stanford University, Stanford, USA
\end{enumerate}

\bigskip

\noindent \textbf{†}  Kailai Sun and Yuebing Liang contributed equally. \\
\noindent \textbf{*} To whom correspondence should be addressed: Shenhao Wang and Jinhua Zhao. E-mail: shenhaowang@ufl.edu;jinhua@mit.edu.
\noindent Yuebing Liang and Yunhan Zheng contributed to this research during this postdoctoral stint at SMART, Singapore.
\bigskip

\vfill


\newcommand{\quickwordcount}[1]{%
  \immediate\write18{texcount -1 -sum -merge -q #1.tex output.bbl > #1-words.sum }%
  \input{#1-words.sum} words%
}
\newcommand{\detailtexcount}[1]{%
  \immediate\write18{texcount -merge -sum -q #1.tex output.bbl > #1.wcdetail }%
  \verbatiminput{#1.wcdetail}%
}



\end{titlepage}

\pagebreak

\section{Abstract}
Urban development has been a defining force in human history, shaping cities for centuries. However, past studies mostly analyze such development as predictive tasks, failing to reflect its generative nature. Therefore, this study designs a multimodal generative AI framework to envision sustainable urban development at a global scale. By integrating prompts and geospatial controls, our framework can generate high-fidelity, diverse, and realistic urban satellite imagery across the 500 largest metropolitan areas worldwide. It enables users to specify urban development goals, creating new images that align with them while offering diverse scenarios whose appearance can be controlled with text prompts and geospatial constraints. It also facilitates urban redevelopment practices by learning from the surrounding environment. Beyond visual synthesis, we find that it encodes and interprets latent representations of urban form for global cross-city learning, successfully transferring styles of urban environments across a global spatial network. The latent representations can also enhance downstream prediction tasks such as carbon emission prediction. Further, human expert evaluation confirms that our generated urban images are comparable to real urban images. Overall, this study presents innovative approaches for accelerated urban planning and supports scenario-based planning processes for worldwide cities.

\newpage
\section{Introduction}
The world has undergone significant urban development over the past several centuries. Urban residents comprise 55\% of the global population, a figure projected to rise to 68\% by 2050 \cite{UN2018Urbanization}. The pace of global urbanization has accelerated in the 21\textsuperscript{st} century and is expected to continue increasing in the foreseeable future. Urban development is often characterized by high population and building densities, which foster urban agglomeration effects \cite{melo_meta-analysis_2009}, linked to higher levels of economic development, productivity, innovation, and social capital \cite{glaeser_wealth_2009, xia_analyzing_2020}. However, this rapid urbanization has also brought significant sustainability challenges, including increased climate risks, urban heat island effects, heightened vulnerability to flooding and urban sprawl~\cite{barrington-leigh_century_2015}.

To study and understand the impact of urban development, satellite imagery has become an essential tool for monitoring, evaluation, and prediction because it provides extensive spatial and temporal coverage. Additionally, researchers have widely adopted deep learning techniques to extract key urban metrics from satellite imagery, providing a comprehensive understanding of the past and present trends of global cities. These metrics include urban density indicators, such as building footprints, built-up height \cite{zhou2022satellite}, and land use patterns \cite{ frolking2024global}. Beyond physical structures, deep learning has also been used to infer socioeconomic attributes from satellite imagery, enabling the analysis of population density, poverty levels \cite{jean_combining_2016}, social inequality \cite{thacker2019infrastructure}, greenery exposure, building energy \cite{yap2025revealing}, and flood risk \cite{liu_high-spatiotemporal-resolution_2020, tellman_satellite_2021, boo_high-resolution_2022, wu_improved_2023}. By integrating satellite imagery with deep learning, these studies have greatly enhanced the ability to monitor and analyze global cities, particularly in the data-sparse developing countries, reinforcing its role as a critical tool in global urban research.

However, urban development is not just about analyzing past and present trends—it is also a generative process in which governments, planners, and citizens actively envision and shape future cities. While predictive AI has excelled in analyzing urban patterns, researchers have rarely investigated how to proactively envision urban development using Generative AI (GenAI) \cite{Li2025DisruptiveAI, zheng_urban_2025}. Unlike traditional predictive models, GenAI can synthesize realistic and diverse urban imagery and layout guided by text inputs, offering an unprecedented opportunity to reimagine urban development through generative visualization. 
In urban contexts, GenAI techniques, primarily Generative Adversarial Networks (GANs), have been used for synthesizing building footprints, transferring map styles, and designing spatial plans \cite{zheng_spatial_2023}. More recently, diffusion models have been employed to generate street scenes \cite{dubey_ai-generated_2024, zhuang_hearing_2024}. However, current applications remain limited by data availability, technological capabilities, and analytical scope. Beyond traditional spatial layouts, modern urban planning has also become deeply complex and interdisciplinary, requiring simultaneous reasoning about environmental sustainability \cite{collins2024making}, social equity \cite{xu2025using}, infrastructure accessibility \cite{zheng_urban_2025} and 15-minute city \cite{abbiasov202415}. Yet despite these advances, existing GenAI approaches for urban planning remain fundamentally constrained. First, current frameworks lack comprehensive controllable mechanisms for handling the inherent complexity of urban designs and planning, including quantitative planning indicators, geospatial structures, and free-form multimodal contextual inputs, limiting their ability to translate planning concepts into spatially coherent urban forms. Second, current frameworks are typically trained on limited regions and fail to generalize across diverse global urban patterns, restricting the global scalability for worldwide urban studies. Third, the effectiveness of synthetic urban imagery for enhancing downstream scientific tasks remains insufficiently unexplored.

This study proposes to envision global urbanization through generative visualization in a multimodal generative AI framework guided by language prompts and environmental constraints. Our framework is a domain-adapted variant of Stable Diffusion, fine-tuned for urban imagery to capture urbanization processes by integrating quantitative density indicators, geospatial structures and free-form textual inputs. The framework is applied to 500 metropolitan areas worldwide, covering a diverse range of urban developmental stages \cite{pan_urban_2013}. Empirical results demonstrate that our framework generates satellite imagery that is not only realistic and diverse but also precisely aligned with textual descriptions and environmental constraints (e.g., road networks or elevation maps). The framework effectively envisions the same geographic location under different urban development stages, producing imagery that accurately reflects specified urban density metrics. Beyond its capacity to generate context-aware satellite imagery, our framework also enables interpretable cross-city transfer in the latent space, leveraging spatial constraints from a source area to generate alternative satellite imagery for a target area. Additionally, our findings also reveal that the framework can enhance downstream prediction tasks (e.g., carbon emission prediction). These capabilities have far-reaching implications for urban designers, planners, and policymakers, offering an efficient, creative, and adaptable platform for cross-cultural policy learning, scenario testing, and accelerated future urban planning.

\section{Results}

We introduce a multimodal diffusion framework (Extended Data Fig.1, see Methods for details) to generate realistic and controllable urban satellite imagery conditioned on textual, spatial, and numerical inputs, such as urban density metrics, digital elevation maps, and road networks. The framework incorporates a numeral-aware text encoder for quantitative density metrics alignment, a pseudo-Siamese Mamba network for geospatial constraints, and modules for counterfactual synthesis, local inpainting-based urban redevelopment. 

It supports cross-city transfer across 500 cities, capturing diverse urban-form patterns and learning an interpretable and distinguishable latent representation for each city.  Additionally, by generating synthetic satellite data as a training resource for data augmentation, our framework can enhance global fossil fuel carbon emission prediction. 
 
\begin{figure*}[ht!]
\centering
\includegraphics[width=\textwidth]{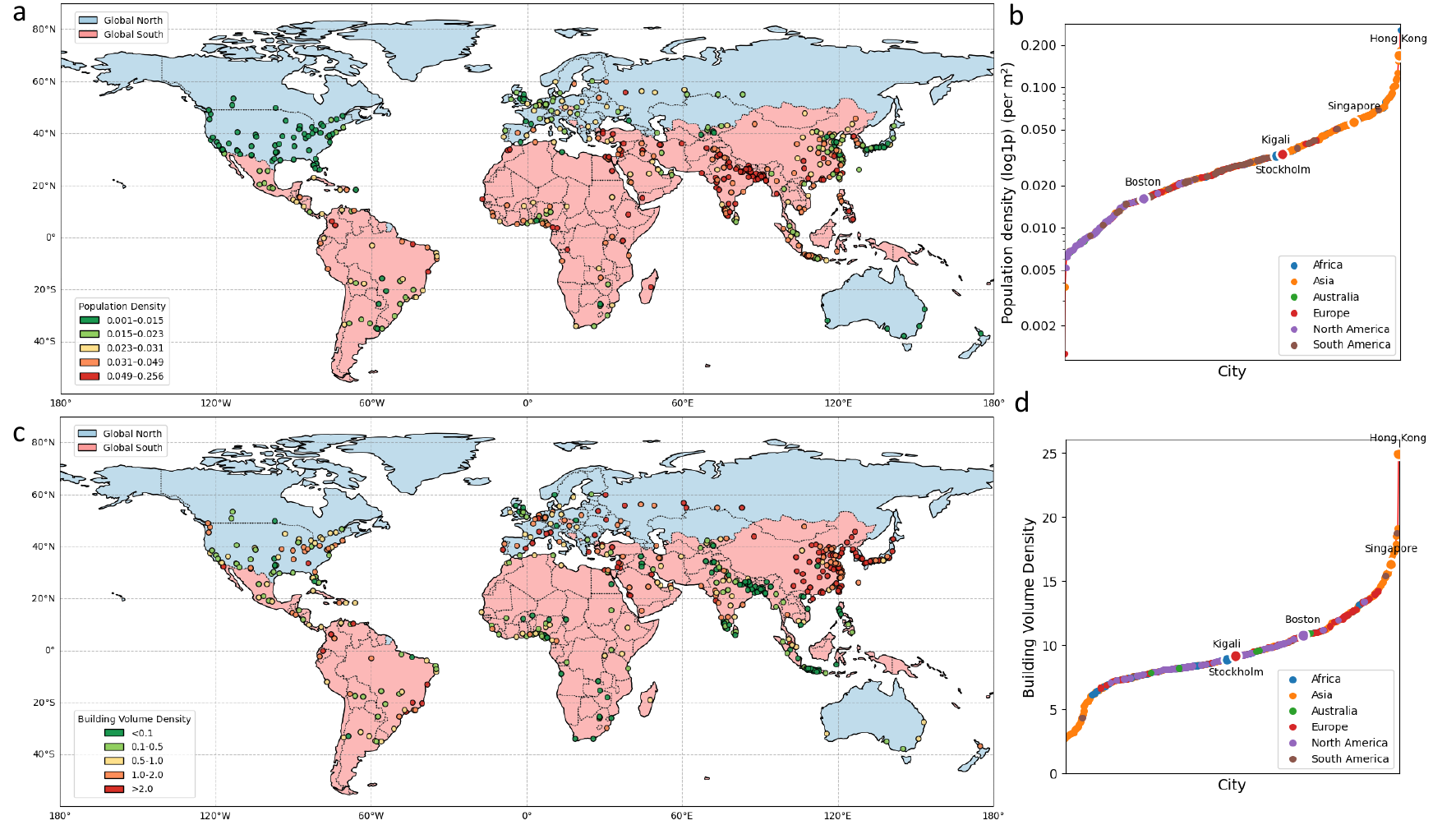}
\caption{\textbf{Spatial heterogeneity of population density and building volume density for global 500 cities.} \small{\textbf{a.} City-level global population density. \textbf{b.} Distribution of population density across 500 cities, sorted in ascending order. \textbf{c.} City-level global building volume density. \textbf{d.} Distribution of building volume density across 500 cities, sorted in ascending order.  In \textbf{a} and \textbf{c}, the city-level densities are divided into five qualitative levels, where warm (orange and red) and cool colors (green and slight green) refer to high and low densities, respectively. In \textbf{b} and \textbf{d}, each dot corresponds to a city and is color-coded by continent (Europe, Asia, North America, South America, Oceania, and Africa). Representative cities, including Boston, Stockholm, Kigali, Singapore, and Hong Kong, are highlighted for reference.}}
\label{fig1} 
\end{figure*}

Our study examines 500 cities across 117 countries on six continents, encompassing diverse urbanization levels and geographical contexts (Fig.\ref{fig1}a-d). To ensure consistency in comparing global cities, we define a standardised spatial unit—a 400m × 400m grid cell—approximately the size of a walkable neighbourhood block \cite{Weng_Ding_Li_Jin_Xiao_He_Su_2019,Moreno_Allam_Chabaud_Gall_Pratlong_2021}. We sample approximately 2,000 grid cells per city, yielding a total of 1 million data samples for training and testing. At each grid cell, we extract satellite imagery from the Mapbox API to analyze urban development. The degree of urban development is described by satellite imagery and further by population and building metrics, such as residential population density, building volume density, and land coverage. Land use ratios across key categories (e.g., residential, commercial, green space) are also included to capture functional diversity.

Urban development stages are quantified through two dimensions: population and buildings. Population density is measured using (1) Residential Population Density (RPD)—the number of residents per unit of residential land area—and (2) Residential Volume Per Capita (RVC), calculated as the total residential building volume divided by the population. Building density is assessed using (3) Building Volume Density (BVD)—total building volume per unit of land area—and (4) Building Coverage Ratio (BCR), the proportion of land covered by buildings. (5) Land Use Ratios (LUR) are also considered, including many categories: residential, commercial, industrial, public facilities, and transportation infrastructure. Green areas (e.g., urban parks, tree-covered spaces, and recreational land) are incorporated to capture the ecological and recreational functions of urban space. Each land use type is spatially encoded and used to analyze the functional heterogeneity of urban grids. These metrics provide a comprehensive and multidimensional view of urban form and development. We compute these metrics for each grid using high-resolution global data from the Global Human Settlement Layer (GHSL) \cite{pesaresi2024advances} (see Methods for details).


\subsection{Generating compliant, realistic, and diverse urban imagery}
\begin{figure*}[!htbp]
\centering
\includegraphics[width=\textwidth]{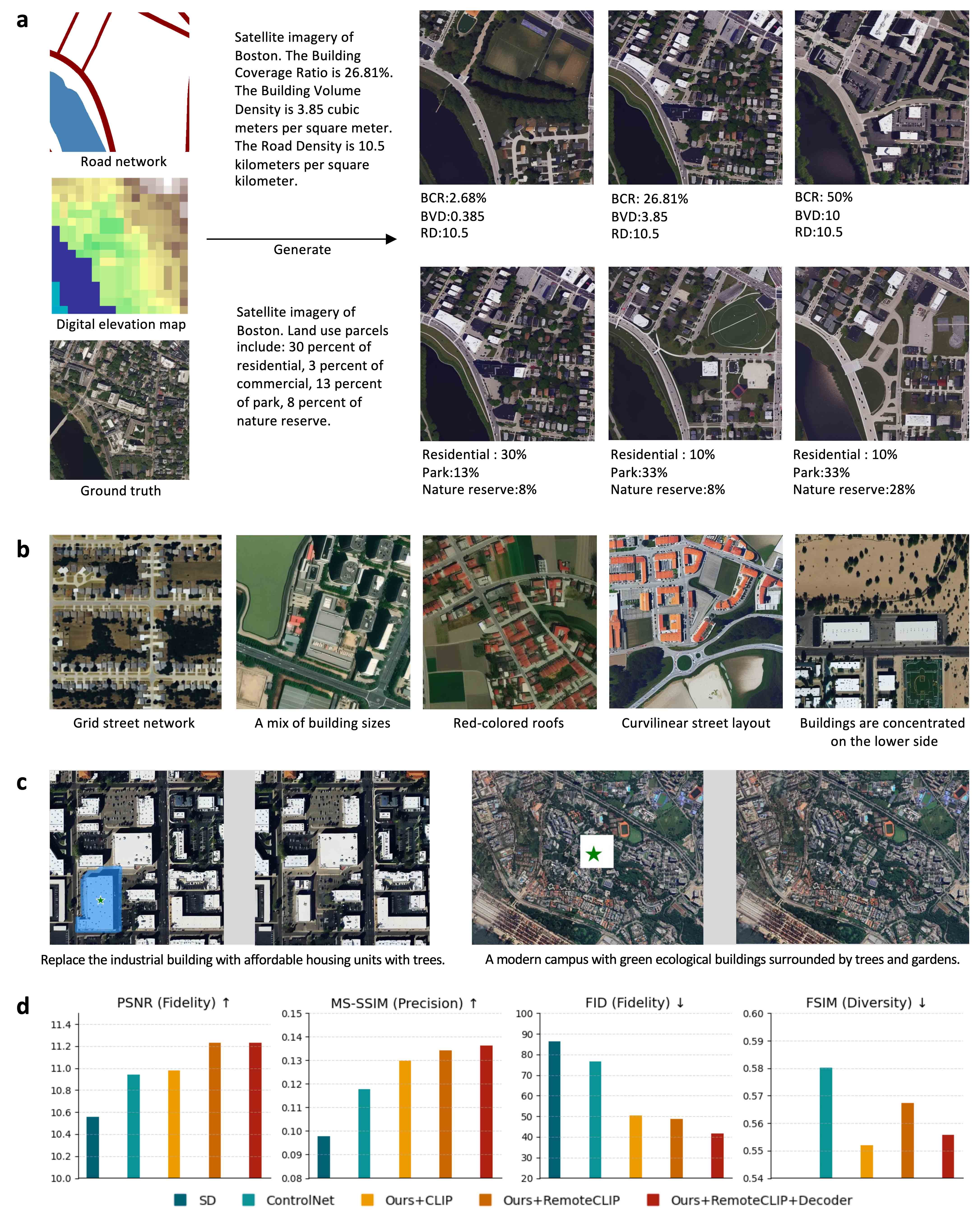}
\vspace{-1cm}
\caption{\small{\textbf{Multi-modal conditional counterfactual synthesis of urban satellite imagery.}  \textbf{a}.  Generated images respond to counterfactual template texts. The first row shows three generated images from the density metric prompt, and the second row shows three generated images from the land use prompt. \textbf{b}. Generated images respond to free-form texts. \textbf{c}. Urban redevelopment with the image inpaint model. \textbf{d}. Generative fidelity, precision and diversity performance of our framework.}} 
\label{fig_counterfactual} 
\end{figure*}

Using simple text prompts that encode city density metrics, our multimodal GenAI framework creates synthetic satellite images that closely align with the existing urban form and density patterns, even for counterfactual scenarios where no real satellite imagery exists (Fig.\ref{fig_counterfactual} and Supplementary Fig.2). 
Synthetic urban imagery can respond to text prompts and environmental constraints (e.g., road networks and elevation maps) across four dimensions: urban density, land use, free-form text, and urban redevelopment. We find that the synthetic satellite image depicts more buildings (Fig.\ref{fig_counterfactual}a) if we increase the building coverage rate and building volume density in the prompt. Conversely, the synthetic satellite image depicts fewer buildings when we decrease the density value in the prompt. It is noteworthy that in Fig.\ref{fig_counterfactual}a and Supplementary Fig.5a, we observe that the building height increases and building type changes (from residential buildings to big commercial blocks) in generated images when the ratio of BVD to BCR (i.e., BVD/BCR) increases in the prompt. Another example lies in Extended Data Fig.2a (Kigali, Africa), where an increase in building density in the prompt leads to the emergence of more industrial constructions in generated images. More examples can be found at Extended Data Figs.2 and 3 and Supplementary Fig.3a. 

On the other hand, we find that the generated satellite image responds sensitively to counterfactual land-use ratios. In the second row of Fig.\ref{fig_counterfactual}a, when the residential ratio decreases (from 30\% to 10\%) and the park ratio increases (from 13\% to 33\%), the generated image reveals spatial patterns consistent with the changes in land-use ratios. This results in a relatively high green access index (GAI is 0.562), a strong clustering tendency (vegetation aggregation index increases from 1.039 to 1.111) and a walkable road intersection density (RID increases from 0.563 to 0.625). 
Similarly, when the residential ratio decreases (from 30\% to 10\%) and the nature reserve ratio increases (from 8\% to 28\%), the generated image exhibits a more walkable appearance (RID achieves 0.94) and similar GAI (0.453).  Another example in Extended Data Fig.2a (Kigali, Africa) shows that increasing the industrial land-use ratio in the prompt leads to more industrial facilities and urban development. More examples can be found at Extended Data Figs.2 and 3. These suggest that our framework has effectively captured different land-use patterns across global cities.

Adding free-form textual modifiers for features such as roof color and the spatial arrangement of urban elements leads to visually plausible correlates in the generated images. In Fig.\ref{fig_counterfactual}b and Supplementary Fig.3b, we observe that our framework can effectively respond to users' free-form textual inputs, including diverse roof colors (e.g., red, blue, write), street layouts (e.g., grid, curvilinear),  building location (e.g., upper, bottom, left, right), and so on, achieving flexible and fine-grained visual control.

Apart from the urban designs in the fixed-scale image or overall image, our framework can also envision local sub-regional urban designs by integrating vision-language inpainting models. On the left of Fig.\ref{fig_counterfactual}c, with users' prompt, our framework can change the sub-regional industrial building into new houses and trees in Phoenix city. Besides, our framework can redevelop the local sub-region by learning from the surrounding environments. On the right of Fig.\ref{fig_counterfactual}c, the new university region in Singapore is designed and redeveloped by prompting green ecological buildings and gardens. In summary, the four dimensions collectively enable the controllable generative ability of our framework, ensuring both visual realism and spatial–semantic consistency across global cities.


\subsection{Generative fidelity, precision and diversity.}
Ideally, synthetic satellite images should closely match the distribution of the real images in terms of fidelity and precision, while capturing the full range of variability present in real-world imagery (diversity).

Our proposed framework outperforms baselines (e.g., Stable Diffusion and ControlNet) in all metrics (Fig.\ref{fig_counterfactual}d and Table S1), achieving the best FID (41.5),  PSNR (11.23), precision FSIM (0.566) and precision LPIPS (0.544). Specifically, we find: (1) Even with limited fine-tuning (20k steps), our framework improves fidelity and precision over the baseline, while extended training (60k steps) further boosts performance (Table S1). (2) Switching the default general-domain CLIP text encoder with domain-specific text encoders (e.g., RemoteCLIP \cite{remoteclip} in Fig.\ref{fig_counterfactual}d) slightly improves model performance without degrading visual quality (e.g., reduce FID by 6.5\% and LPIPS by 3.3\%, increase PSNR by 5.9\%). (3) Additional decoder fine-tuning yields the strongest overall performance across fidelity and precision metrics, reducing FID by 14.6\% and diversity-SSIM by 3.22\%, while further improving precision FSIM and LPIPS. These findings indicate that our framework is accurate and robust for high-quality satellite image generation.




To evaluate how well synthetic satellite images conveyed density-relevant information, a ResNet50-based regression model trained on real satellite images was used to estimate synthetic satellite images' BCR and BVD (Table S2), using the density labels of the in-distribution test set as ground truth. The assumption holds that the text prompt represented the same information between real and synthetic satellite images. The Mean Absolute Error (MAE) and R-Square were used to evaluate the performance (Table S2). The lower MAE and higher R-Square indicate better regression performance. The performance (R-Square 0.871 in BVD, 0.619 in BCR) suggests that the synthetic satellite images can obey the density condition in prompts.

Beyond the evaluation via computer vision metrics, it is valuable to conduct a reader study as a human evaluation for assessing synthetic urban images. In Fig.\ref{fig3}c, for fair evaluation, we randomly shuffle the ground-truth images or generated urban images, and the experts were blinded as to which image was real or generated (see Methods for details). Overall, we find that the generated urban images are comparable to those of real urban images. For example, in the first sub-figure of Fig.\ref{fig3}c, we evaluated our framework's generation performance in the urban density dimension. We find that the generated urban images achieve similar performance with real images in four aspects: site constraints satisfaction, realism, density alignment and road network connectivity. We did not find any statistical significance (P>0.05) for the score difference between real and generated urban images by t test. Similarly, in the urban land use dimension, there is no statistical significance for the score difference in two aspects: site constraints satisfaction and road network connectivity. The significance (P=0.008) in the land use aspect demonstrates that our generated images align better with land use ratios in the prompt than real images, because generated images show relatively distinct building outlines. Similarly, we also did not find any statistically significant score difference between urban redevelopment and original development. Furthermore, we also ask experts: "Is our AI tool useful for you or the urban domain? (1-10 score)" The experts assigned our framework a high utility score of 8.22/10, underscoring its potential as a practical tool for the urban planning community.

\FloatBarrier
\subsection{Cross-city learning through generative AI}
Urbanization always involves cross-city learning in design, planning, or policy-making. In fact, the developing countries often follow the urbanization steps of the developed countries, in terms of institutional building, environmental design, and economic policies. Such a learning process could be conveyed through generative visualization. Here we present the generative visualization using the status quo of a source region as the environmental constraint and another region as the target environment for learning. Such source-target region pairs are applied to Hong Kong-Singapore, Lusaka-Boston.

\begin{figure*}[!htbp]
\centering
\includegraphics[width=\textwidth]{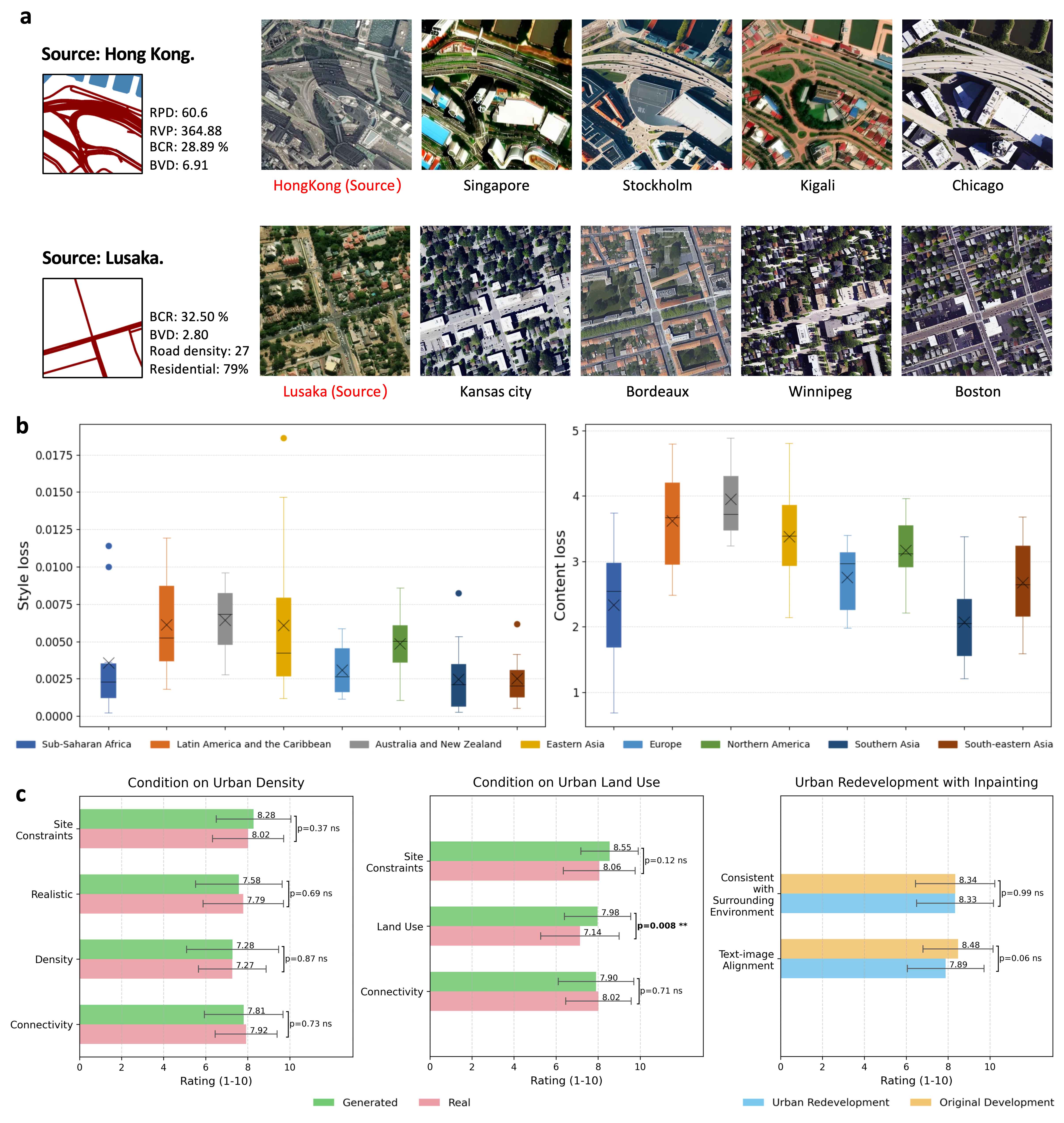}
\vspace{-1cm}
\caption{\small{\textbf{Cross-city learning through generative urban visualization.} \textbf{a}. Generated satellite images across cities from the source city (e.g., Hong Kong, Lusaka). \textbf{b}. Performance variation across regions. \textbf{c}. Human evaluation of our framework by urban experts. 42 urban experts (from the United States, Singapore, China, etc.; from both academia and industry) were recruited to rate (1-10 scores) for 136 questions. The generated urban images are comparable to those of real urban images.}}
\label{fig3} 
\end{figure*}

In Fig.\ref{fig3}b, we evaluate the transferability of our framework across different cities. Style loss and content loss are used to evaluate the performance of city stylization (Section S5).  We compute the city style/content loss between synthetic satellite images and the original real-world dataset for each city. Then, the averaged style loss achieves 0.176 and the content loss achieves 2.912. Our framework achieves lower values in both averaged losses than the baseline (style loss is 0.242 and content loss is 3.105), demonstrating enhanced transfer quality and more effective city stylization performance. 

In Fig.\ref{fig3}b, among all regions, Sub-Saharan Africa, South-eastern and Southern Asia demonstrate the lowest median style losses with narrow interquartile ranges (IQRs), indicating strong intra-regional stylistic consistency which is effectively learned by our framework. In contrast, Latin America and the Caribbean, Eastern Asia and Australia and New Zealand exhibit the higher median style loss with broader IQRs, suggesting that cities in these regions challenge our framework with more heterogeneous and diverse visual features. Besides, Europe and Northern America yield relatively low and stable style losses, which may be attributable to the regularity and standardization of urban forms. Besides, we also analyzed content loss to evaluate how well the spatial structure and semantic layout of the target cities were preserved during generation. Latin America and the Caribbean, Eastern Asia and Australia and New Zealand exhibited consistently higher content loss, reflecting their complex urban forms with dense, multi-scale spatial layouts. While Southern Asia and Sub-Saharan Africa demonstrated the lowest content losses, implying that the spatial organization is more standardized, and thus easier for our framework to generate.

\subsection{Latent vectors capture urban visual cross-city learning patterns.}

\begin{figure*}[!htbp]
\centering
\includegraphics[width=\textwidth]{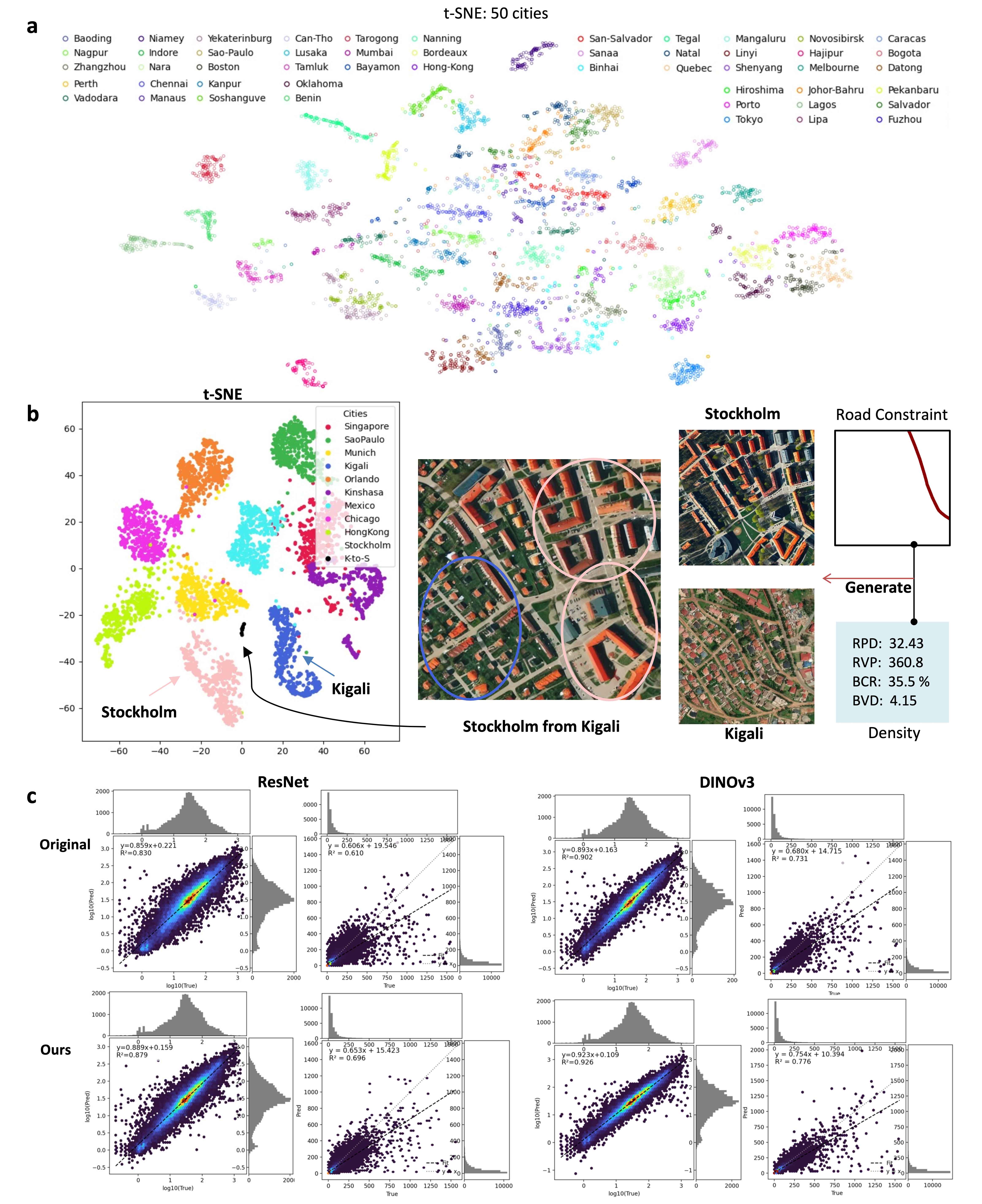}
\caption{\small{\textbf{Our framework quantifies urban visuals in latent space.} \textbf{a}. Our framework has learned a distinguishable visual representation for each city. other city's results can be found in SI. \textbf{b}. Cross-city transfer via prompt modification from “Stockholm” to “Stockholm from Kigali” yields hybrid satellite imagery, with latent features (black points) located between Kigali and Stockholm clusters. \textbf{c}. Performance comparison of predicted versus true global fossil fuel carbon emission for ResNet and DINOv3, using original data (top) and generative augmentation (bottom).}} 
\label{fig_urban_latent_space} 
\end{figure*}

To further analyse the learned visual patterns across different cities by our framework, the latent feature dimension is reduced by principal component analysis (PCA) and t-distributed stochastic neighbor embedding (t-SNE) in Figs.\ref{fig_urban_latent_space}a. The visual features of cities are distinguishable, demonstrating that our framework has learned the distinguishable visual representation for each city. Our findings align with the unique urban form and style of each city. For example, from Fig.\ref{fig_urban_latent_space}b, among the 10 example cities shown, clear grouping patterns emerge. Our findings reveal the proximity between Munich and Stockholm, indicating that the model captured their common European urban structures. Chicago and Orlando, as North American cities, show moderate separation from other clusters while still exhibiting internal compactness, reflecting consistent intra-city visual styles. In contrast, Kinshasa and Kigali, representing more irregular urban patterns typical of African cities, form separate but nearby clusters, possibly reflecting shared morphological traits (e.g., low development patterns). 

We observe that the learned latent patterns can interpret urban visual cross-city learning. For example, when we perform a cross-city transfer by modifying the prompt from “Stockholm” to “Stockholm from Kigali”, the generated satellite imagery exhibits a hybrid morphology, retaining Stockholm’s grid-like layout while reflecting Kigali’s architectural texture and vegetation patterns. Interestingly, the corresponding latent features (black points) are positioned between the clusters of Kigali and Stockholm. In Fig.\ref{fig_urban_latent_space}b, while the global layout and building typologies resemble Stockholm (e.g., orange roofs, grid-like planning), localized patterns such as plot sizes and vegetation distributions retain visual traits from Kigali. For human evaluation, we also conduct a reader study by recruiting 42 urban experts to determine whether the generated images exhibit a hybrid morphology of both source cities. 37 experts correctly identified the successful cases, achieving an 88\% consensus rate. These observations show that our framework can blend stylistic and structural features from both the source and target cities, reflecting a learned disentanglement between spatial layout and city-specific styles.

\subsection{Enhancing global fossil fuel carbon emission prediction}

A use case for our synthetic satellite image is the training of deep learning models for downstream tasks (e.g., global carbon emission prediction). To investigate the value of using synthetic images as training data, a CNN model (ResNet), a Transformer (Swin Transformer), and a recent vision foundation model (DINOv3) were finetuned to predict global fossil fuel carbon emissions in global 111 cities (1k samples per city). 

Robust performance improvement is observed with generative data augmentation (Fig.\ref{fig_urban_latent_space}c, Table S3 and Supplementary Fig.6). We find that all models exhibit significant increases in the coefficient of determination (R\textsuperscript{2}) and reductions in mean absolute error (MAE) and root mean square error (RMSE). Specifically, for ResNet34, we observe that the R\textsuperscript{2} increased from 0.610 to 0.696 (R\textsuperscript{2}\textsubscript{log} from 0.830 to 0.879), and the MAE decreased from 22.04 to 18.42. For DINOv3 (SAT-493M), which benefits from large-scale satellite image pretraining, R\textsuperscript{2} improved from 0.731 to 0.776 (R\textsuperscript{2}\textsubscript{log} from 0.902 to 0.926), and MAE fell from 17.22 to 14.65, surpassing all baselines without or with other data augmentation methods \cite{cubuk2020randaugment, xiao2025sample}. Beyond overall accuracy, both ResNet and DINOv3 exhibit tighter clustering around the 1:1 line when trained with synthetic imagery (Fig.\ref{fig_urban_latent_space}c), indicating reduced bias in both low- and high-emission regimes. This suggests that synthetic imagery provides complementary visual diversity that stabilizes carbon emission regression performance under long-tailed scenarios, consistent with recent computer vision studies showing that synthetic data can enhance model generalization performance \cite{ma2025fully, trabucco2023effective}. These findings demonstrate that synthetic imagery effectively enriches the semantic and spatial diversity, allowing models to better capture fine-grained contextual features of fossil fuel activity.

Spatial heterogeneity in prediction performance was also observed (Supplementary Fig.6). The divergence largely reflects regional differences in fossil fuel dependence and energy-use patterns. Predictions in rapidly urbanizing and less developed cities were more accurate because fossil fuel combustion remains the dominant source of energy for industry, transport, and power generation, showing a strong spatial relationship between visible urban form (e.g., dense road networks, industrial zones) and CO\textsubscript{2} emissions. However, developed cities exhibit lower prediction accuracy. Their energy systems are more diversified, integrating renewables, nuclear, and district heating, resulting in a weak relationship with visible urban form.

\section{Discussion}

Urban development is an essential process across all global cities, promising good access to economic growth, environment, transportation, and sustainability, for achieving the United Nations' Sustainable Development Goal 11. Recent evidence reveals that satellite imagery is an effective tool for monitoring (e.g., flood risk \cite{tellman_satellite_2021}, human greenspace exposure \cite{wu_improved_2023}), prediction (e.g., poverty levels \cite{jean_combining_2016}, building energy \cite{yap2025revealing}), and interpretation \cite{abitbol2020interpretable}.  As noted earlier, however, urban development is not just about analyzing past and present trends—it is also a generative process where governments, planners, and citizens actively envision and shape future cities. GenAI is expected to assist human planners in the urban layout generation process \cite{zheng_urban_2025,Li2025DisruptiveAI}. Compared to GANs-based urban applications (e.g., building footprint synthesis and spatial layout generation \cite{zheng_spatial_2023, quan_urban-gan_2022}), diffusion models provide more reliable performance and control over high-quality image generation, making them well-suited for sustainable urban scene synthesis \cite{dubey_ai-generated_2024, zhuang_hearing_2024}. 

Our findings build on prior research highlighting the substantial contribution of generative multimodal AI models for global urban development. Specifically, our study introduces a powerful framework to generate effective urban alternatives with a controllable development process at the micro level and to envision urban development across global cities at the macro level. We show that this multimodal generative AI framework maps urban planning descriptions of textual prompts by human planners to visual urban designs. Due to the geospatial constraints of the built environment, our framework enables controllable urban image synthesis by combining natural language (e.g., by prompting for variations in urban density) with spatial inputs (e.g., road network layouts or elevation maps). Trained on a global dataset, our framework has the generalisation ability to support the envisioning of the counterfactual urban development (e.g., counterfactual building metric and land use in Fig.\ref{fig_counterfactual}a, Extended Data Figs.2 and 3) and fine-grained control by conditioning on free-form text prompts (Fig.\ref{fig_counterfactual}b and Supplementary Fig.3b). Inpainting model's results (Fig.\ref{fig_counterfactual}c) reveal that it can envision local sub-regional urban designs and urban redevelopment by learning from the surrounding environments and planners' inputs. These specific findings further highlight the significant flexibility and potential of diffusion models in urban planning.


Our framework can render an accurate and diverse urban satellite imagery beyond prior generative methods. Experiment (Table S1) reveals that switching the default CLIP text encoder with a domain-specific text encoder (i.e., RemoteCLIP \cite{remoteclip}) can bring an improved performance. One explanation lies in that the RemoteCLIP has been trained with a large-scale text-image remote sensing dataset, aligning the vision-language representations with rich semantics of satellite imagery. Our experiments (Table S1) also reveal that the LongCLIP text encoder with longer (248) tokens leads to better improvements in SSIM and MS-SSIM, and results in reduced diversity. A possible explanation is that LongCLIP better memorizes the input prompts, thus generating visually consistent but less diverse images. 
More importantly, the effectiveness of our GenAI framework is verified by human expert evaluation.



Urban development inherently involves cross-city learning, as cities around the world often look to one another for guidance in urban development, environmental design, and policy implementation. Our findings reveal that the framework has learned a distinguishable visual representation for each city (Fig.\ref{fig_urban_latent_space}a). Thus, it supports the potential of urban development in assisting global planners and designers by finetuning across the global 500 cities' data. In particular, developing regions frequently draw inspiration from the urbanization experiences of more developed areas. Our findings reveal that the framework can provide a practical approach to this process by supporting cross-city generative design (Fig.~\ref{fig3}). By conditioning image generation on the environmental constraints of a source city and the textual description of a target city, the framework can produce synthetic satellite imagery that blends structural features and stylistic elements from both locations (Kigali learns from Stockholm in Fig.\ref{fig_urban_latent_space}b). This reveals that the framework has learned diverse urban styles and form patterns from the global dataset, demonstrating the generalization ability to disentangle and recombine visual and spatial features across geographies. Beyond the synthetic visual data, the framework's structured latent representation, extracted by PCA and TSNE, provides interpretable visual embeddings that distinguish city-specific style and urban form patterns, serving as a foundation for downstream analytical tasks such as city clustering, socioeconomics analysis, and cross-city comparisons.


Our results provide evidence and offer a powerful framework to support government agencies, urban planners, and private developers in envisioning global urban development, conducting comparative analyses, performing city style transfers, and exploring urban scenarios across diverse global contexts, especially in developing countries, to achieve sustainable development goals. This study reveals that large GenAI models can support the creation and generation of urban designs by transforming planners’ prompts, which introduce key urban planning concepts and constraints, into detailed visual images (e.g., layouts), guided by user-defined draft plans. Our framework demonstrates the potential of generative AI to accelerate the urban planning process. Our analysis of cross-city learning, urban latent representation and counterfactual urban development, will help cities achieve better and sustainable urban development strategies for city equality and economic growth. For the downstream applications (e.g., global fossil fuel carbon emission prediction), our findings support the potential of synthetic data in augmenting real training data to achieve better prediction performance (Fig.\ref{fig_urban_latent_space}, Table S3 and Supplementary Fig.6). By releasing our framework as open access, we encourage a more democratic planning process, enabling not only professional planners and policymakers but also citizens and communities to adapt and apply it toward customized urban design and planning that aligns with SDGs.

Our GenAI framework can establish a transformative paradigm for global earth observation by extending its scientific value to geosciences, environmental science, social economics, and computer science. By facilitating counterfactual urbanization scenarios across 500 cities in building density, population, and land-use ratios, it provides a cross-disciplinary simulator where geoscientists can simulate climate risks from satellite images, ecologists can assess how counterfactual urban density (e.g., population growth) impacts biodiversity and explore the dynamic evolution of green space boundaries, social scientists can visualize policy outcomes at a global scale and analyse how the urbanization affect economic development and urban equity, and computer scientists can explore how to improve the numerical sensitivity (e.g., urban metric) of LLMs, accelerating scientific discoveries and technical advancements in alignment with United Nations' SDGs.


Although our framework demonstrates the potential in urban development, several limitations remain. First, the long-term urban development data for each city should be included. This time-series data can present trends of urban evolution. But collecting long-term data (e.g., 1975-2025 years) in underdeveloped and developing countries (e.g., Africa) is challenging. Second, further alignment with human intent using technologies (e.g., reinforcement learning from human feedback \cite{ouyang2022training}) may improve the generated quality for specific domain requirements. For example, because urban planning workflows differ across cities, future work may encode these differences and local government polices as configurable process templates to facilitate interactive and human-in-the-loop urban planning, ensuring the tool aligns with real practice. Another future research direction is to explore urban style loss and content loss as urban predictive metrics. We apply style loss and content loss to evaluate our model and assess visual differences between cities (Fig.\ref{fig3} and Section S5). Higher style loss indicates greater architectural diversity and cultural influences, while lower style loss suggests design uniformity. It may correlate with cultural heritage or historical preservation. Higher content loss may be linked to rapid urbanization or mixed-use zoning, with irregular layouts showing more content loss compared to grid-based cities like New York. In the future, style and content loss with specific improvements will have the potential to serve as predictive metrics for understanding different development features (e.g., substantial vs. style). High content loss with low style loss could suggest substantial growth and style preservation. Tracking these losses over time helps assess how cities balance historical preservation with modernization, and how they address density management and urban sprawl.

\section{Methods}
\textbf{Datasets} 

\vspace{1mm}
\noindent \textbf{Global urban boundaries.} We select 500 cities worldwide as our study sites. Urban boundaries are derived from the GHS Urban Centre Database \cite{maririvero2024ghsurban}, which defines urban areas as clusters of 1km × 1km grid cells with a population density of $\geq$ 1,500 inhabitants/$km^2$ and a total population of $\geq$ 500,000 inhabitants. To ensure sufficient data representation of each city's geographical context, we include only metropolitan areas with a land area exceeding 100 $km^2$. For consistency in global comparisons, we subdivide each urban area into 400m × 400m grid cells and sample 2,000 grid cells per city, yielding a dataset of 1 million samples. We randomly allocate 80\% of the spatial units within each metropolitan area to the training set and reserve the remaining 20\% for testing. 

\vspace{1mm}
\noindent \textbf{Satellite imagery.} 
We use Mapbox imagery data\footnote{\url{https://docs.mapbox.com/api/maps/static-tiles/}}, which provides publicly accessible satellite and aerial imagery as raster tilesets at varying zoom levels for global cities. These tilesets are sourced from a combination of commercial providers and open dataset, such as NASA and USGS. To ensure precise alignment with 400m × 400m grid cells, we download high-resolution satellite tiles (<0.78m/pixel), then crop and merge them to match each grid’s spatial extent. The final images are 512 × 512 pixels, providing a resolution of ~0.78m/pixel.

\vspace{1mm}
\noindent \textbf{Population and building density metrics.} We utilize data from the Global Human Settlement Layer (GHSL) \cite{pesaresi2024advances} to compute density metrics. The latest release, GHSL P2023A, provides global population and building density datasets at 100m resolution for 2020. Specifically, GHS-BUILT-S \cite{pesaresi2023ghsbuits} reports total and non-residential built-up surface area per grid cell, derived from Sentinel-2 composite and Landsat imagery. GHS-BUILT-V \cite{pesaresi2023ghsbuiltv} provides total and non-residential built-up volume, estimated through a joint assessment of Sentinel-2, Landsat, and global DEM data. GHS-POP \cite{carioli2023ghspop} reports population counts derived from CIESIN GPWv4.11 \cite{ciesin2018gpwv4}. Empirical evidence suggests that GHSL is among the most reliable publicly available datasets for estimating global urban density \cite{pesaresi2024advances}. Using these datasets, we compute four density metrics  for each 400m × 400m grid cell: (1) Residential population density (RPD) = number of people / residential built-up surface area. (2) Residential volume per capita (RVC) = residential built-up volume / number of people. (3) Building volume density (BVD) = total built-up volume / land area. (4) Building coverage ratio (BCR) = total built-up surface area / land area. 


\vspace{1mm}
\noindent \textbf{Environmental constraints.} We collect environmental constraints from OpenStreetMap (OSM)\footnote{\url{www.openstreetmap.org}}, a public dataset that provides vectorized representations of urban features. Specifically, we extract water bodies, railway infrastructure, and major roads (i.e., motorways, trunks, primary, secondary, and tertiary roads as classified in OSM), excluding smaller roads to minimize their influence on the generation of urban patterns. To ensure spatial alignment, we perform a spatial intersection between the vectorized environmental data and each grid cell. The intersected layers are then rasterized into 512 × 512 pixel images, serving as image controls for our framework.

\begin{figure}
\centering
\includegraphics[width=0.99\linewidth]{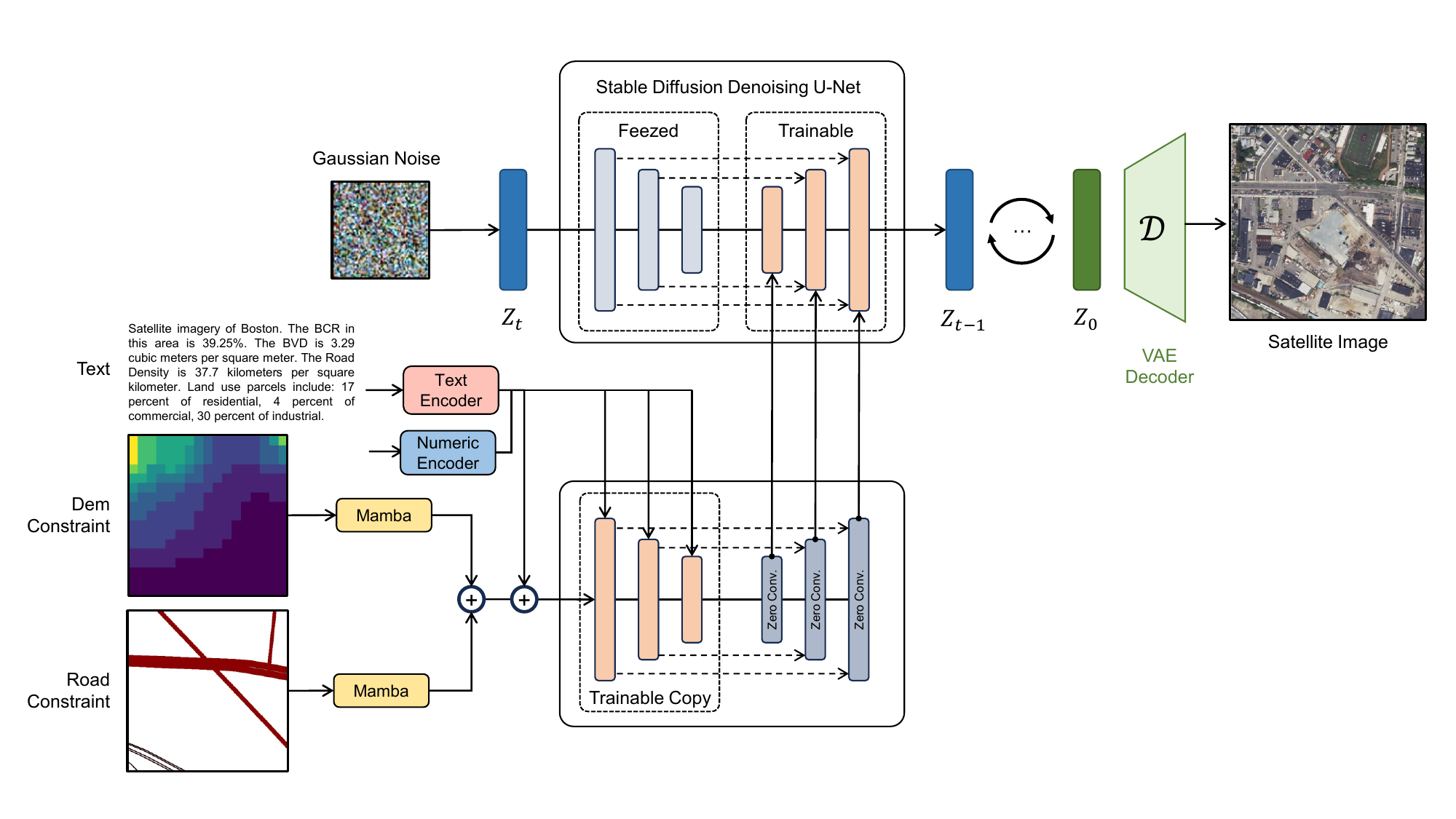}
\caption*{\textbf{Extended Data Fig. 1} Modeling diagram of our framework.}
\label{fig:UCN}
\end{figure}

\vspace{1mm}
\noindent \textbf{Our framework}\\
Our framework is a modified version of diffusion models designed for precise control over image synthesis through structural conditioning (Extended Data Fig.1). It is a multi-modal diffusion framework for satellite image generation, which integrates text descriptions (e.g., land use, building coverage) and environmental constraints (elevation and road maps).  First, the input text provides detailed urban information (e.g., "Satellite imagery of Boston. The BCR in this area is 39.25\%. The BVD is 3.29 m³/m². The Road Density is 37.7 km/km². Land use parcels include: 17\% residential, 4\% commercial, 30\% industrial" ). These semantic descriptions are encoded through a text encoder. Meanwhile, digital elevation map and road maps are processed and fused together by Mamba-based deep models. All multi-modal information is injected into a trainable copy of the SD encoder, and the fused features are integrated into the main Stable Diffusion model via zero convolutions. The model iteratively denoises latent noise into a latent vector, which is then decoded by a VAE decoder to produce the final satellite image.

Given a dataset of corresponding environmental constraint map, DEM map, text and satellite image pairs $
\mathcal{D} = \left\{ \left(c_{\text{constraint}},c_{\text{dem}},c_{\text{text}}, x_{\text{pixel}}\right)_i \right\}_{i=1}^{n}$, we want to learn the neural network parameter $\theta$ to generate the satellite image $x_{\text{pixel}}$, conditioning on the environmental controls and text input. In the forward diffusion process, the Gaussian noise is added to the latent representation of $x_{\text{pixel}}$ created by encoding the image with the VAE’s encoder component $\mathbf{E}_V$: $\mathbf{x}_{0}=\mathbf{E}_V(x_{\text{pixel}})$. In the reverse diffusion process of SD, for each pair $i \in (1, n)$ and timestep $t \in (0, T)$, the neural network gradually samples the latent vectors $\mathbf{x}_{t-1}$ from the random Gaussian noise $\mathbf{x}_{T}$. The distribution obeys:

\begin{equation}
p_\theta(\mathbf{x}_{t-1} \mid \mathbf{x}_t,\mathbf{c}) = \mathcal{N}(\mathbf{x}_{t-1}; \mu_\theta(\mathbf{x}_t, t,\mathbf{c}), \Sigma_\theta(\mathbf{x}_t, t,\mathbf{c})),
\end{equation}
where conditions $c=f(\mathbf{E}_I( c_{\text{constraint}},c_{\text{dem}}),\mathbf{E}_T (c_{\text{text}}))$. $\mathbf{E}_I$ is the image encoders and $\mathbf{E}_T$ is the text prompt encoder. Then, the mean of this Gaussian distribution obeys:

\begin{equation}
\mu_{\theta}(x_{t}, t,c) \;=\; 
\frac{1}{\sqrt{\alpha_{t}}}\Bigl(
  x_{t} \;-\; 
  \frac{1 - \alpha_{t}}{\sqrt{\,1 - \bar{\alpha}_{t}\,}}\,
  \epsilon_{\theta}(x_{t}, t,c)\Bigr),
\end{equation}
where parameters $\{\bar{\alpha}_{t} \in (0,\,1)\}_{t=1}^{T}$ and $\bar{\alpha}_{t} \;=\; \prod_{i=1}^{t} \alpha_{i}$. SD parameterises the Gaussian noise term instead to make it predict  $\epsilon_{t}$ from the input $x_{t}$ at time step t. At last, A mean squared error loss computed between the true and predicted noises $\epsilon_{t}$ and $\epsilon_{\theta}$ enables to compute gradients and update the weights of the unfrozen components:


\begin{equation}
L \;=\; 
\frac{1}{|D|} \sum_{x_{0}^{i},\,c^{i} \in D} \;\sum_{t=1}^{T} \
\bigl\lVert \epsilon_{t} \;-\; 
\epsilon_{\theta}\bigl(\sqrt{\bar{\alpha}_{t}}\,x_{0}^{i} \;+\; \sqrt{1 - \bar{\alpha}_{t}}\,\epsilon_{t},\,t,\,c^{i}\bigr)
\bigr\rVert^{2}
\end{equation}

\begin{figure}
\centering
\label{fig:twocase}
\includegraphics[width=0.95\linewidth]{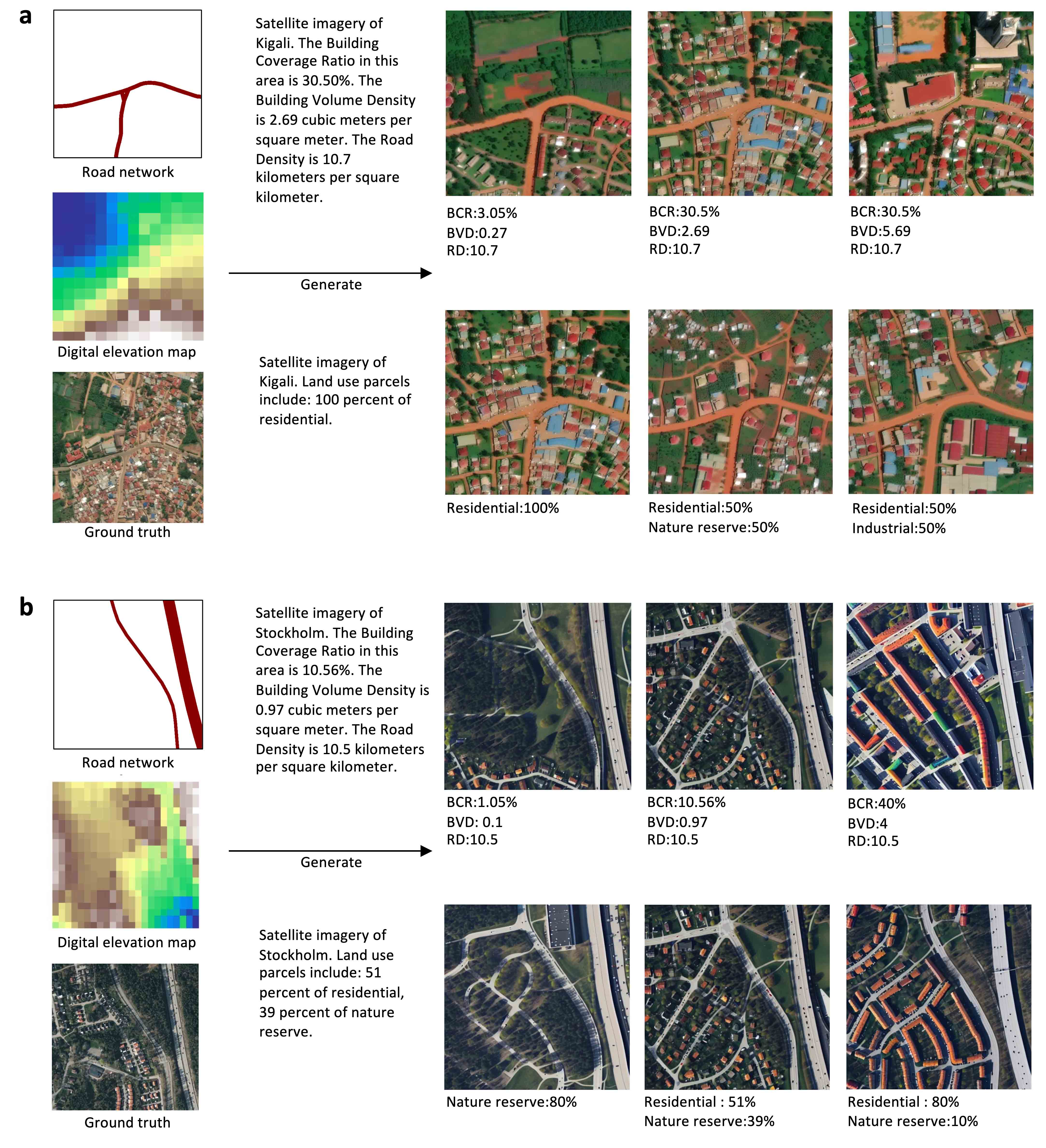}
\caption*{\textbf{Extended Data Fig. 2}  Multi-modal (text-image) conditional counterfactual synthesis of urban satellite imagery in Kigali and Stockholm cities. \textbf{a}.  Generated images respond to counterfactual texts in Kigali City. \textbf{b}. Generated images respond to counterfactual texts in Stockholm City. The first row shows three generated images from the density metric prompt, and the second row shows three generated images from the land use prompt.  }
\end{figure}

\begin{figure}
\centering
\label{fig:twocase}
\includegraphics[width=0.95\linewidth]{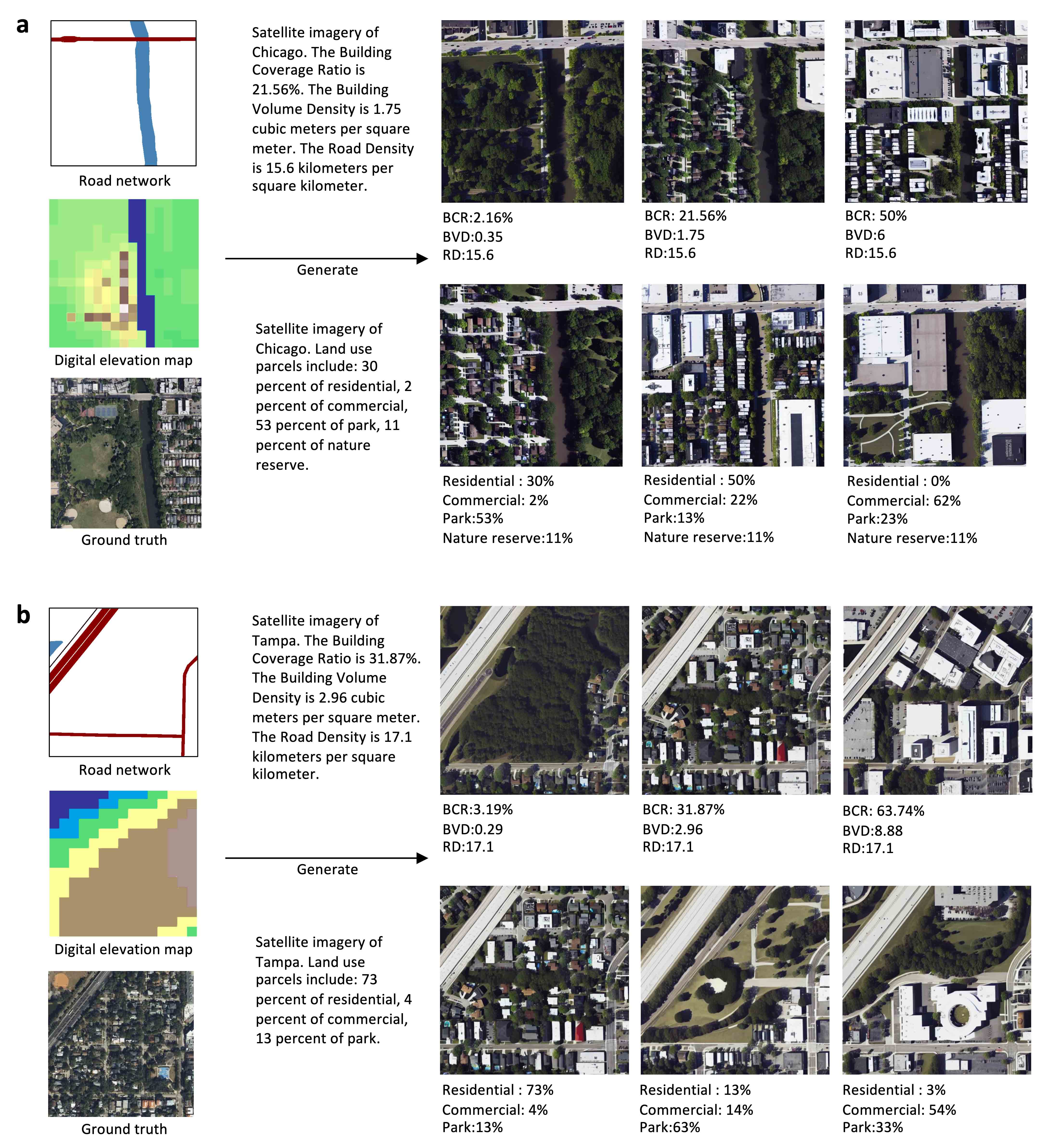}
\caption*{\textbf{Extended Data Fig. 3} Multi-modal (text-image) conditional counterfactual synthesis of urban satellite imagery in Chicago and Tampa cities. \textbf{a}.  Generated images respond to counterfactual texts in Chicago City. \textbf{b}. Generated images respond to counterfactual texts in Tampa City. The first row shows three generated images from the density metric prompt, and the second row shows three generated images from the land use prompt. }
\end{figure}

\begin{figure}
\centering
\label{survey}
\includegraphics[width=0.95\linewidth]{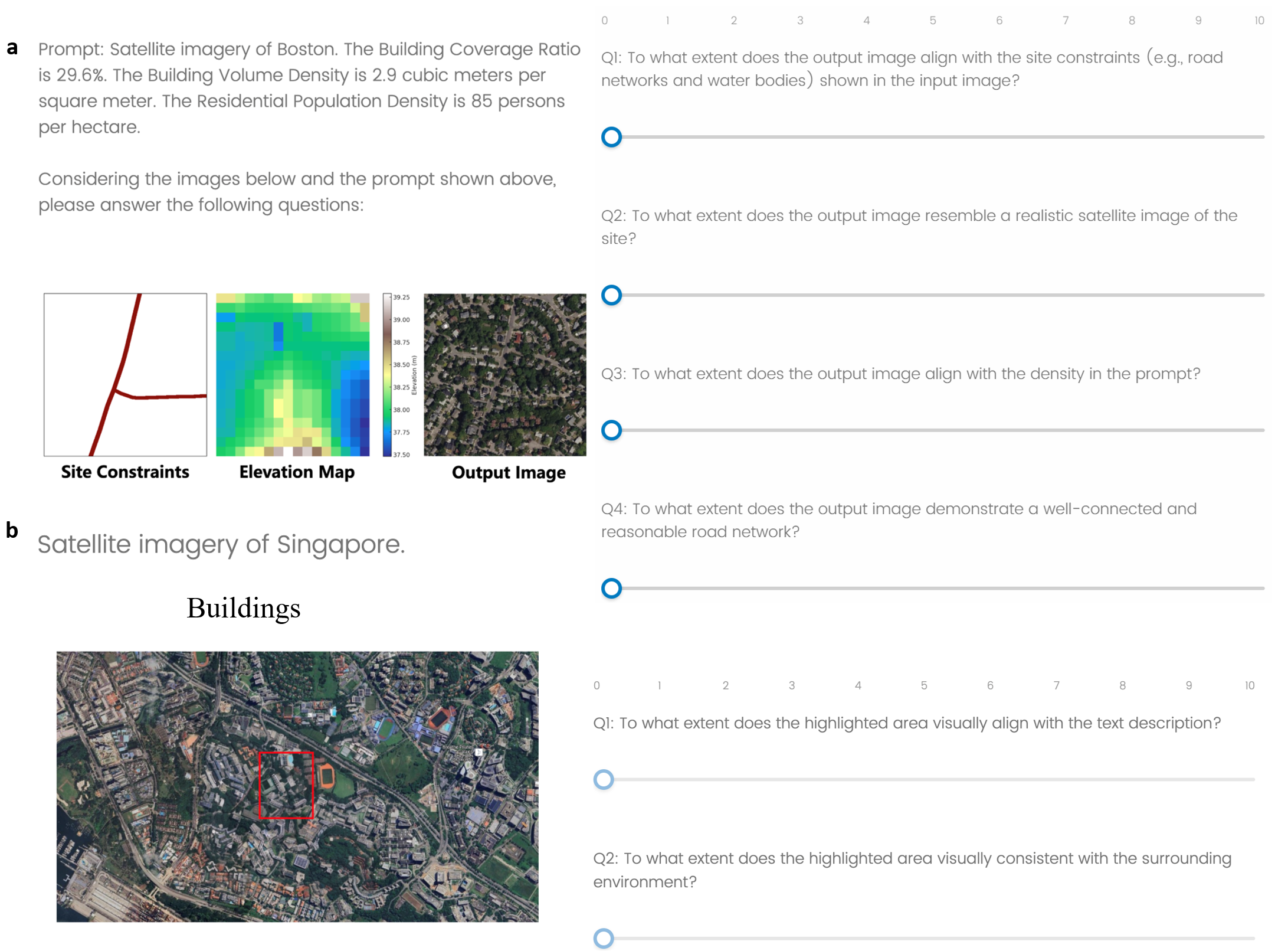}
\caption*{\textbf{Extended Data Fig. 4.} Examples of reader study. \textbf{a}.  An example question for urban experts to evaluate generated urban image in the urban density dimension. \textbf{b}. An example question for urban experts to evaluate generated urban redevelopment image.}
\end{figure}

Recognizing the unique characteristics of urban imagery compared to natural scenes, our framework incorporates several critical enhancements beyond the original ControlNet framework (Extended Data Fig. 1). Our framework builds on ControlNet \cite{zhang_adding_2023}, an extension of diffusion models that enables fine-grained control over image generation by conditioning on structural inputs. Given the distinct nature of urban images compared to natural images, our framework introduces several key innovations over the original ControlNet. Conventional methods primarily condition image generation on captions or images but struggle with numerical constraints, such as density metrics. To address this, we introduce a numeral-aware network to enhance the text embedding, which effectively integrates numerical data as conditioning information (Supplementary Fig.5). Second, existing approaches typically consider single constraint input, which is insufficient to fully capture urban environments from multiple perspectives. To overcome this, our framework incorporates a multi-source pseudo-Siamese Mamba network (Supplementary Fig.4), enabling the joint input of multiple spatially aligned controls (e.g., digital elevation map and road constraint map). 

Besides, to enable the urban image inpainting function, we finetune the Stable Diffusion Inpainting (SDI) model on our datasets. We mask some local sub-regions and finetune SDI to inpaint them for completing the original image, using the StableDiffusionInpaintPipeline function in the Diffusers library. Also, we apply the public inpaining anything model for this study.

\vspace{1mm}
\noindent \textbf{Fine-tuning Hyperparameters.}\\
All fine-tuning experiments were conducted on a high-performance computer with a 64-bit Ubuntu 22.04 system, a 128-core Intel(R) Xeon(R) Platinum 8462Y+ Processor, 2 TB  physical memory (RAM) and 8 NVIDIA H200 graphics processing units (GPUs). our framework is fine-tuned using satellite images at a resolution of 512 × 512 pixels. The batch size is set to 16, and the full 32 floating-point (FP32) precision is applied to ensure the model's performance. To accelerate the training, the Distributed Data Parallel (DDP) strategy ensures efficient and synchronized gradient updates across multiple devices. During training, each GPU holds a copy of the model and processes a different subset of the data, with gradients being synchronized at each backward pass via NCCL-based communication. In this setting, fine-tuning an our framework model takes approximately 5 days for 60k training steps with 6 NVIDIA H200 GPUs. Model weights for ControlNet and standard SD (version 1.5) were obtained from GitHub \footnote{https://github.com/lllyasviel/ControlNet}.

\vspace{1mm}
\noindent \textbf{Baselines.}\\
Many evidences validate that Stable Diffusion is an excellent Text-to-Image (T2I) model.
Besides, previous research proposed adding conditional control (e.g., images) to T2I generative models \cite{zhang_adding_2023} can generate more controllable images. In this study, as a strong baseline, ControlNet is fine-tuned on our dataset and compared with our proposed framework under the same setting.

\vspace{1mm}
\noindent \textbf{Fine-tuning experiments.}\\
For fine-tuning, four experimental dimensions are explored: (1) the effect of switching from the general CLIP text encoder to domain-specific text encoders (e.g., RemoteCLIP) ;(2) the impact of the number of training steps (e.g., 20k vs. 60k); (3) different strategies for fine-tuning components in the SD pipeline, including fine-tuning only the ControlNet or jointly fine-tuning the decoder. 

To address the domain gap between CLIP’s training data and the urban remote sensing context of our study, we replaced the default CLIP encoder with RemoteCLIP \cite{remoteclip}, which is specifically pre-trained on geographic and satellite imagery captions. Besides, the original CLIP encoder limits input prompts to 77 tokens, which is insufficient for our relatively long urban layout descriptions. LongCLIP \cite{zhang2024long}, which supports a longer input limit (248 tokens), was integrated in this fine-tuning experiment. In all experiments, the variational autoencoder (VAE) component was kept frozen, as it has been shown to generalize well without additional tuning.

The original ControlNet architecture \cite{zhang_adding_2023} reuses the Stable Diffusion encoder as a deep and robust backbone to capture diverse control signals, which is validated by some previous studies and evidence \cite{zhao2023unleashing, xu2023open}. Accordingly, ControlNet adopts a trainable copy of the SD encoder during fine-tuning. We follow this standard setup in the first stage of our experiments. In the second stage, however, we further unlock the SD decoder for fine-tuning. This second stage proves particularly beneficial when our large training dataset exhibits domain-specific styles (satellite imagery and urban city style). Similar to DreamBooth, this two-step fine-tuning can be interpreted as simultaneously fine-tuning our framework and performing a task-specific adaptation. In this setting, training setups use a low learning rate ($1e^{-5}$) while fine-tuning the decoder with a lower learning rate ($2e^{-6}$) to stabilize training and prevent overfitting.

\vspace{1mm}
\noindent \textbf{Reader study}\\
42 urban experts (from the United States, Singapore, China, etc.) were recruited to rate (1-10 scores) for 136 questions. We gathered ground-truth images and generated urban images, and then presented them to urban experts, who evaluated them by rating them based on their expertise. We sent questionnaires to global urban experts via MIT's online Qualtrics Survey Platform \footnote{https://ist.mit.edu/qualtrics/survey}. For fair comparison and evaluation, we randomly shuffle the ground-truth images or generated urban images, and the experts were blinded as to which image was real-world or generated. For each image, the responses were rated (1-10 scores) based on the coherence of urban structure, visual realism, overall plausibility, and experts' general judgment and visual interpretation. Two examples can be found at the Extended Data Fig.4. Besides, for free-form generation task, we ask: "Which image reveals an urban environment closer to the language description?" and all experts give right answers.

\vspace{1mm}
\noindent \textbf{Latent feature dimensionality reduction.}\\
For this experiment, to visualize and analyze the internal representations learned by our framework, we extract latent features from the DDPM backbone with an initial shape of $4 \times 64 \times 64$. A spatial average pooling operation is first applied to reduce the spatial resolution to $4 \times 32 \times 32$, compressing local details while retaining high-level semantics. The resulting features are then flattened and projected into a 128-dimensional space using Principal Component Analysis (PCA). Finally, the features are embedded into a two-dimensional space using t-Distributed Stochastic Neighbor Embedding (t-SNE) for visualization purposes.

\vspace{1mm}
\noindent \textbf{Density regression models.}\\
The challenge in this research area lies in accurately evaluating the density of synthetic samples. In Table \ref{tab:main1}, the density metrics of generated satellite images were evaluated using a fine-tuned regression model based on a ResNet50 backbone pre-trained on ImageNet.The model's fully connected layer was modified to output multiple continuous values (i.e., multi-task regression), each representing a specific urban metric (e.g., BCR, BVD). A sigmoid activation was cascaded to the final outputs to normalize predictions within a limited range, matching the scaled ground truth. Prior to training, all input satellite images were resized to 224×224 and normalized using the standard ImageNet statistics (mean = [0.485, 0.456, 0.406], std = [0.229, 0.224, 0.225]). The model was trained for 20 epochs using Smooth L1 loss and the Adam optimizer with a learning rate of 0.0003. Real-world satellite images were only used for training and validation, while the testing was performed on generated satellite images to evaluate the density metrics of the generated satellite images.

\noindent \textbf{Global fossil fuel carbon emission prediction models.}\\
To predict the global fossil fuel carbon emission, we finetune ResNet34, Swin Transformer and DINOv3 on the open-data inventory for anthropogenic carbon dioxide (ODIAC) \footnote{https://odiac.org/index.html}. In the data part, data from 111 cities were used to conduct this experiment (1k samples per city). To match the resolution of ODIAC and our generated images, we apply a weighted average method to interpolate and upsample the resolution (from 1 km x 1 km to 400 m x 400 m). The long-tailed distribution of fossil fuel carbon emissions is transformed into a Gaussian-like distribution using a logarithmic transformation. We split the real images into training and test sets (7:3).  In the model part, we apply ResNet34 and Swin Transformer (swin\_b) pretrained on the ImageNet dataset, while DINOv3 (ViT-7B) is pretrained on the satellite image dataset (SAT-493M).  To make full use of DINOv3, we extract and fuse the features from the last three layers, then cascade them through MLPs to predict fossil fuel carbon emissions. Similarly,  ResNet and Swin  Transformer are directly cascaded with MLPs to predict fossil fuel carbon emissions. In the training strategy part, we first train our framework to generate new satellite images by conditioning the road network and prompting text: "Satellite imagery of Boston. The fossil fuel CO2 emissions in this area are 52.3 tonnes of carbon". These synthetic images are used for training prediction models. For pair comparison, we keep the same test set and add only synthetic satellite images to the training set.  The MSE loss and AdamW optimizer with a learning rate of 0.0003 are used for model training.












\noindent \textbf{Metrics for fidelity, precision and diversity.} 

(Fidelity) The Fréchet Inception Distance (FID) is used to quantify the difference between the feature distributions of real and generated images. It is computed using the 2,048-dimensional activation vectors extracted from the pool3 layer of an Inception V3 model pre-trained on ImageNet. All images are resized to 512 × 512 pixels and normalized. Each of our framework is evaluated using 50,000 real and 50,000 generated satellite images, providing a robust estimate of the FID score. This sampling strategy exceeds the minimum sample size of 10,000 recommended by Heusel et al.\cite{heusel2017gans}, ensuring greater stability and statistical reliability of the results. A higher PSNR suggests that the generated image retains a larger proportion of the signal (i.e., meaningful features) present in the ground-truth image, reflecting improved fidelity.

(Precision) Real and generated satellite images are evaluated using the same set of metrics as in the assessment of generative diversity. However, the interpretation of these metrics differs: higher MS-SSIM (Multiscale Structural Similarity, computed across three color channels), FSIM, and SSIM scores indicate greater structural similarity between generated images and their corresponding real counterparts, conditioned on the same text prompt and constraint map. In contrast, a lower LPIPS score implies smaller perceptual differences as judged by human vision models, indicating more realistic and perceptually accurate satellite image synthesis.

(Diversity) Generative diversity was assessed by calculating pairwise SSIM and FSIM scores between generated satellite images and their corresponding ground truth. In this context, lower SSIM and FSIM  scores indicate lower structural similarity between images generated from the same text prompt and constraint map, suggesting higher visual diversity within the generated set.

SSIM, PSNR, LPIPS, FSIM, and MS-SSIM were computed using the Pyiqa library \footnote{https://github.com/chaofengc/IQA-PyTorch}, which provides efficient GPU-accelerated implementations optimized for PyTorch. All computations were performed on a machine equipped with 6 NVIDIA H200 GPUs for model deployment, feature extraction, and metric calculation.

\vspace{2mm}
\section*{Data availability} 
Urban boundary data are obtained from the Global Human Settlement (GHS) Urban Centre Database 2023 at \url{https://human-settlement.emergency.copernicus.eu/ghs_ucdb_2024.php}, which provides harmonized 400 m × 400 m grid delineations for 500 metropolitan areas worldwide. Satellite imagery is retrieved through the Mapbox Static Tiles API at \url{https://docs.mapbox.com/api/maps/static-tiles/}. Population and built-up metrics are derived from the GHSL P2023A (2020) suite, including GHS-BUILT-S (\url{https://developers.google.com/earth-engine/datasets/catalog/JRC_GHSL_P2023A_GHS_BUILT_S}), GHS-BUILT-V (\url{https://human-settlement.emergency.copernicus.eu/ghs_buV2023.php}), and GHS-POP (\url{https://human-settlement.emergency.copernicus.eu/ghs_pop2019.php}). Environmental constraints, such as major roads, railways, and water bodies, are extracted from OpenStreetMap at \url{https://www.openstreetmap.org}.

We make most cities' datasets open at \url{https://huggingface.co/datasets/skl24/500city/tree/main}, where each city's data is processed and completed.

\vspace{2mm}
\section*{Code availability} 
The code used for conducting the analysis is accessible on GitHub at \url{https://github.com/kailaisun/UrbanControlNet/}. The trained weights and checkpoints are accessible on Hugging Face at:\url{https://huggingface.co/skl24/Urbancontrolnet/tree/main}.

\bibliography{mybib}

\section*{Acknowledgements}
This research is supported by the National Research Foundation (NRF), Prime Minister’s Office, Singapore under its Campus for Research Excellence and Technological Enterprise (CREATE) programme. The Mens, Manus, and Machina (M3S) is an interdisciplinary research group (IRG) of the Singapore MIT Alliance for Research and Technology (SMART) centre.


\section*{Competing Interests Statement}
The authors declare no competing interests.

\renewcommand{\thepage}{S\arabic{page}}
\renewcommand{\thesection}{S\arabic{section}}
\renewcommand{\thetable}{S\arabic{table}}
\renewcommand{\thefigure}{S\arabic{figure}}

\setcounter{page}{1}
\setcounter{section}{0}
\setcounter{table}{0}
\setcounter{figure}{0}

\begin{titlepage}
{\noindent\LARGE\bf Supplementary Information: Envisioning global urban development with satellite imagery and generative AI}

\bigskip

\begin{flushleft}\large
    Kailai Sun \textsuperscript{1,{\dag}},
    Yuebing Liang \textsuperscript{2,{\dag}},
    Mingyi He \textsuperscript{3},
    Yunhan Zheng \textsuperscript{4},
    Alok Prakash \textsuperscript{1},
    Shenhao Wang \textsuperscript{1,5,{*}},
    Jinhua Zhao \textsuperscript{3,{*}},
    Alex ``Sandy'' Pentland \textsuperscript{6}
\end{flushleft}

\begin{enumerate}[label=\textbf{\arabic*}]
\item Singapore–MIT Alliance for Research and Technology Centre (SMART), Singapore
\item Department of Urban Planning, Tsinghua University, Beijing, China
\item Department of Urban Studies and Planning, Massachusetts Institute of Technology, Cambridge, MA, USA
\item College of Urban and Environmental Sciences, Peking University, Beijing, China
\item Department of Urban and Regional Planning, University of Florida, Gainesville, FL, USA
\item Stanford Institute for Human-Centered Artificial Intelligence, Stanford University, Stanford, USA
\end{enumerate}


\bigskip


\noindent \textbf{†}  Kailai Sun and Yuebing Liang contributed equally. \\
\noindent \textbf{*} To whom correspondence should be addressed: Shenhao Wang and Jinhua Zhao. E-mail: shenhaowang@ufl.edu;jinhua@mit.edu.

\noindent Yuebing Liang and Yunhan Zheng contributed to this research during this postdoctoral stint at SMART, Singapore.
\vfill


\end{titlepage}

\tableofcontents

\newpage
\section{Relationship between Density Metrics}

\begin{figure}[!htbp]
    \centering
    \includegraphics[width=0.6\linewidth]{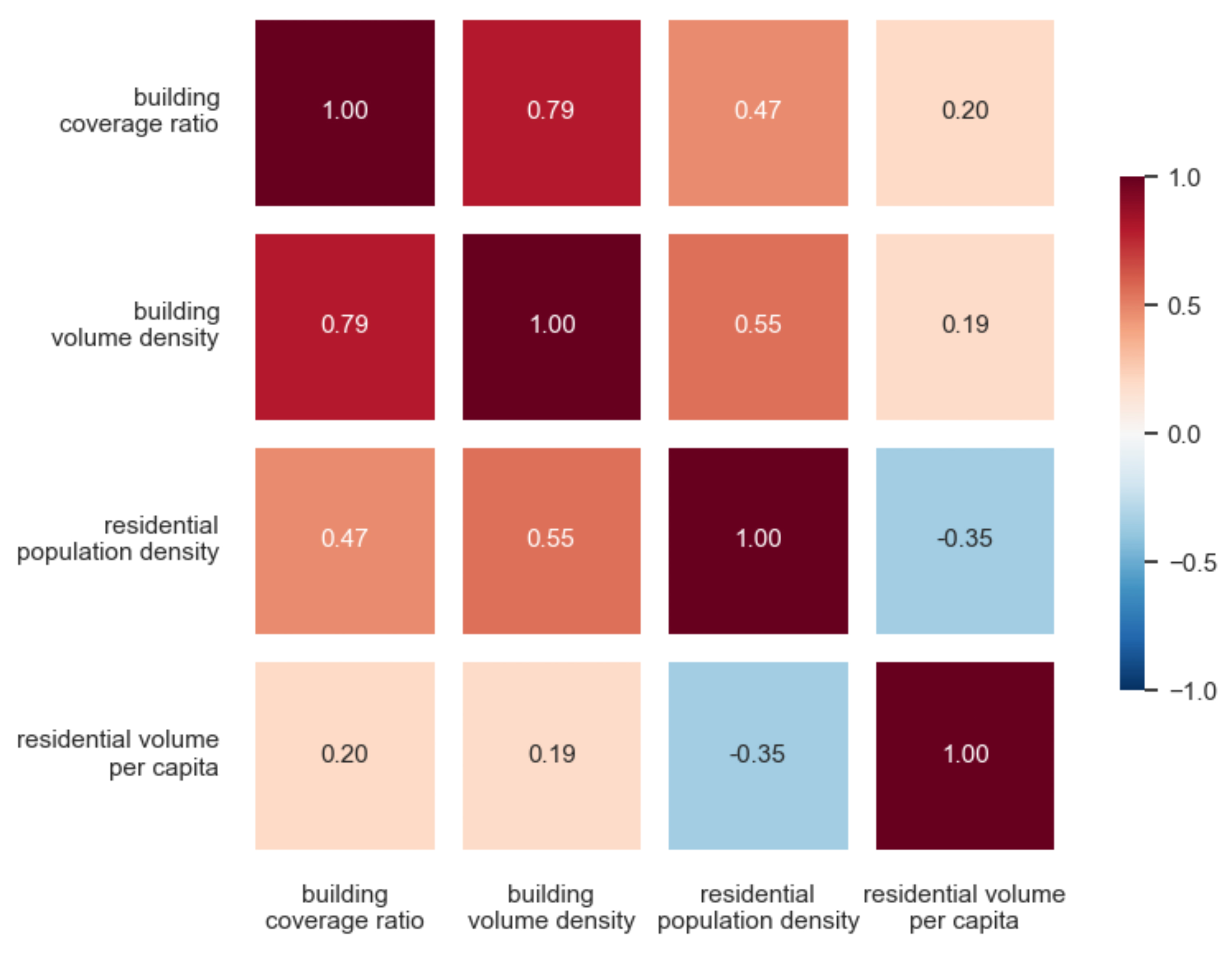}
    \caption{Relationship between Density Metrics}
    \label{fig:SI-1}
\end{figure}

Figure~\ref{fig:SI-1} illustrates correlations between density metrics. There is a strong relationship between building volume density and building coverage ratio (r=0.79), indicating that areas with higher land coverage by buildings also tend to have greater total building volume. Higher building coverage is generally associated with higher population density (r=0.47). Further, residential population density correlates positively with building volume density (r=0.55), suggesting that densely populated areas generally feature higher building volumes. However, higher population density is generally associated with lower residential volume available per person (r=-0.35).

\newpage

\newpage
\section{Generated urban imagery responds to prompts and constraints.}

\begin{figure*}[!htbp]
\centering
\includegraphics[width=\textwidth]{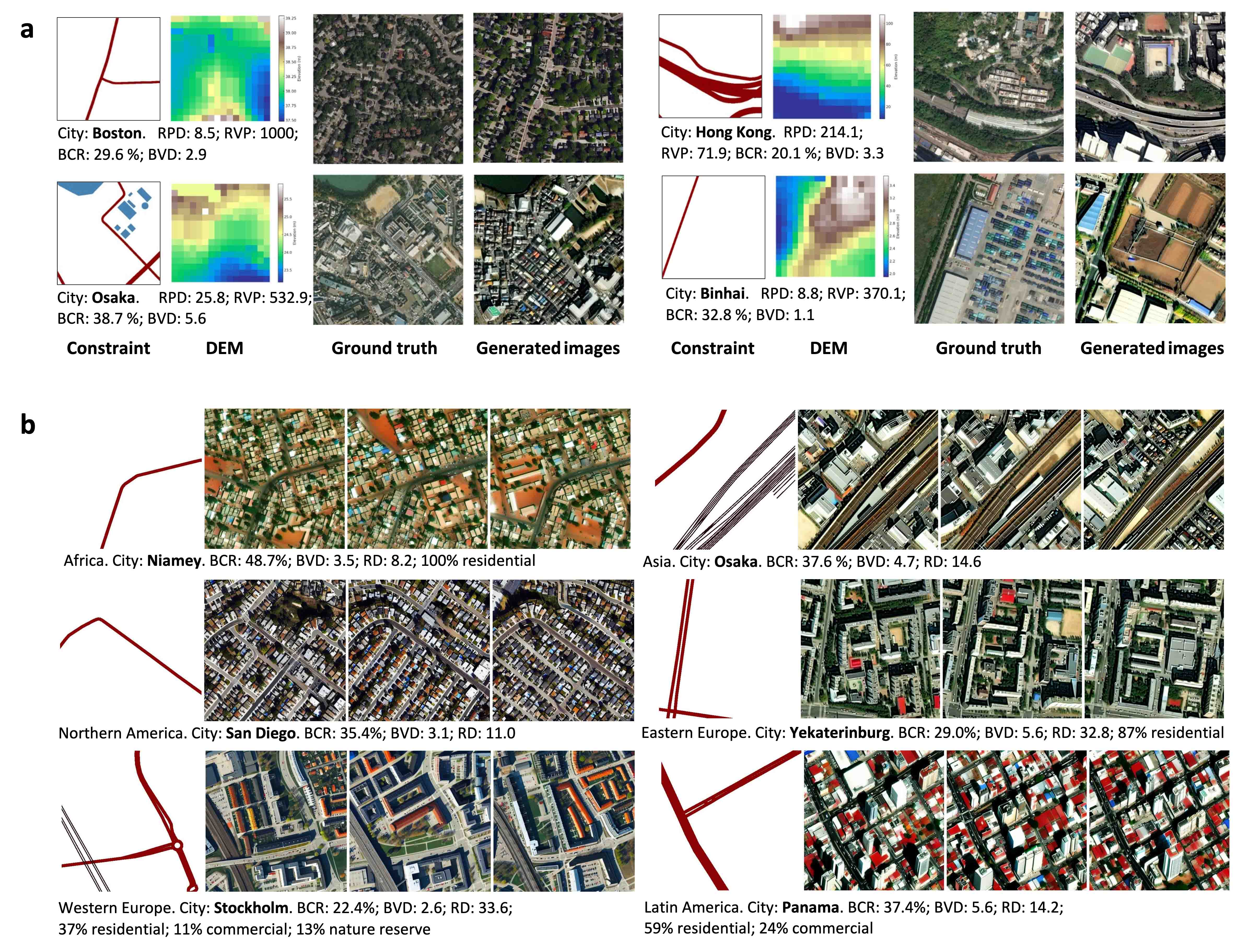}
\caption{\small{\textbf{Generated urban imagery responds to prompts and constraints.} \textbf{a}. Generated images respond to different environmental constraints and text \textbf{b}.  Generated imagery is diverse under the same control. RPD: residential population density; RVC: residential volume per capita; BVD: building volume density; BCR: building coverage ratio; Land use ratios: residential, commercial, industrial, etc. }}
\label{fig_responsive_imagery}
\end{figure*}

Using simple text prompts based on the city density metrics, we find that our our framework can create synthetic satellite images that replicate the existing urban development level with a non-existing satellite imagery (Fig.\ref{fig_responsive_imagery}). Generative urban imagery can respond to text prompts and environmental constraints. In the top-left example (Boston), the generated image closely resembles the ground truth. In the top-right example (Hong Kong), the model moves buildings toward the bottom area, which is approximately 80 meters lower than the top region, according to the digital elevation map (DEM). In the bottom-left example (Osaka), the generated image respects the spatial constraints effectively.
In the bottom-right example (Binhai, Tianjin), the model shifts buildings to the lower-elevation region in the DEM, showing an adaptive response to terrain.

\section{Generative fidelity, precision and diversity.}

\begin{table}[ht]
\small
\setlength{\tabcolsep}{2pt}
\centering
\caption{Quantitative assessment of image fidelity, diversity, and precision metrics}
\label{tab:ucn-compact}
\begin{tabular}{@{}lllcccccccc@{}}
\toprule
\textbf{Models} & \textbf{Steps} & \textbf{Params} &
\multicolumn{2}{c}{\textbf{Fidelity}} &
\multicolumn{2}{c}{\textbf{Diversity}} &
\multicolumn{4}{c}{\textbf{Precision}} \\
& & & FID$\downarrow$ & PSNR$\uparrow$ & SSIM$\downarrow$ & FSIM$\downarrow$ &
SSIM$\uparrow$ & FSIM$\uparrow$ & LPIPS$\downarrow$ & MS-SSIM$\uparrow$ \\
\midrule
\multicolumn{11}{l}{\textbf{Baselines}} \\
 \makecell[l]{Original SD \\ (No finetuning)}
  & 0 & 1.4B & 86.5 & \makecell{$10.56$ \\ \scriptsize{$\pm 0.9$}} & -- & -- &
\makecell{$0.0927$ \\ \scriptsize{$\pm 0.008$}} & \makecell{$0.5556$ \\ \scriptsize{$\pm 0.024$}} &
\makecell{$0.7162$ \\ \scriptsize{$\pm 0.045$}} & \makecell{$0.0978$ \\ \scriptsize{$\pm 0.007$}} \\
ControlNet & 60k & 1.4B & 76.5 & \makecell{$10.94$ \\ \scriptsize{$\pm 2.2$}} & \makecell{$0.1847$ \\ \scriptsize{$\pm 0.13$}}  &  \makecell{$0.5802$ \\ \scriptsize{$\pm 0.06$}}  &
\makecell{$0.1087$ \\ \scriptsize{$\pm 0.07$}} & \makecell{$0.5603$ \\ \scriptsize{$\pm 0.04$}} &
\makecell{$0.5623$ \\ \scriptsize{$\pm 0.038$}} & \makecell{$0.1179$ \\ \scriptsize{$\pm 0.044$}} \\
\midrule
\multicolumn{11}{l}{\textbf{Few Steps}} \\
UCN+CLIP & 20k & 1.5B & 72.5 & \makecell{$10.58$ \\ \scriptsize{$\pm 2.3$}} & \makecell{$0.1665$ \\ \scriptsize{$\pm 0.08$}}  &  \makecell{$0.5560$ \\ \scriptsize{$\pm 0.04$}}  &
\makecell{$0.1235$ \\ \scriptsize{$\pm 0.06$}} & \makecell{$0.5545$ \\ \scriptsize{$\pm 0.035$}} &
\makecell{$0.5808$ \\ \scriptsize{$\pm 0.041$}} & \makecell{$0.1120$ \\ \scriptsize{$\pm 0.048$}} \\
\midrule
\multicolumn{11}{l}{\textbf{Text Encoders}} \\
UCN+CLIP & 60k & 1.5B & \textbf{50.3} & \makecell{$10.98$ \\ \scriptsize{$\pm 2.4$}} &
\makecell{\textbf{0.1645} \\ \scriptsize{$\pm 0.06$}}& \makecell{\textbf{0.5520} \\ \scriptsize{$\pm 0.04$}} & \makecell{$0.1357$ \\ \scriptsize{$\pm 0.05$}} &
\makecell{$0.5576$ \\ \scriptsize{$\pm 0.034$}} & \makecell{$0.5560$ \\ \scriptsize{$\pm 0.038$}} &
\makecell{$0.1299$ \\ \scriptsize{$\pm 0.043$}} \\
UCN+RemoteCLIP & 60k & 1.5B & \textbf{48.6} & \makecell{\textbf{11.23} \\ \scriptsize{$\pm 2.5$}} &
\makecell{$0.1831 $ \\ \scriptsize{$\pm 0.08$}} & \makecell{$0.5673 $ \\ \scriptsize{$\pm 0.05$}} & \makecell{$0.1424$ \\ \scriptsize{$\pm 0.06$}} &
\makecell{\textbf{0.5630} \\ \scriptsize{$\pm 0.039$}} & \makecell{\textbf{0.5476} \\ \scriptsize{$\pm 0.038$}} &
\makecell{$0.1341$ \\ \scriptsize{$\pm 0.018$}} \\
UCN+LongCLIP & 60k & 1.8B & 61.0 & \makecell{$10.97$ \\ \scriptsize{$\pm 2.7$}} &
\makecell{0.2094 \\ \scriptsize{$\pm 0.09$}}
& \makecell{0.5551 \\ \scriptsize{$\pm 0.05$}} & \makecell{\textbf{0.1511} \\ \scriptsize{$\pm 0.06$}} &
\makecell{$0.5502$ \\ \scriptsize{$\pm 0.040$}} & \makecell{$0.5792$ \\ \scriptsize{$\pm 0.049$}} &
\makecell{\textbf{0.1411} \\ \scriptsize{$\pm 0.055$}} \\
\midrule
\multicolumn{11}{l}{\textbf{SD Decoder Fine-tuning}} \\
UCN+CLIP & 20k & 1.5B & 44.4 & \makecell{$10.60$ \\ \scriptsize{$\pm 2.17$}} &
\makecell{\textbf{0.1600} \\ \scriptsize{$\pm 0.06$}}
& \makecell{\textbf{0.5558} \\ \scriptsize{$\pm 0.04$}} & \makecell{$0.1317$ \\ \scriptsize{$\pm 0.05$}} &
\makecell{$0.5571$ \\ \scriptsize{$\pm 0.034$}} & \makecell{$0.5624$ \\ \scriptsize{$\pm 0.04$}} &
\makecell{$0.1265$ \\ \scriptsize{$\pm 0.042$}} \\
UCN+RemoteCLIP & 20k & 1.5B & \textbf{41.5} & \makecell{\textbf{11.23} \\ \scriptsize{$\pm 2.4$}} &
\makecell{0.1772 \\ \scriptsize{$\pm 0.07$}}
 &  \makecell{0.5696 \\ \scriptsize{$\pm 0.04$}}& \makecell{$0.1406$ \\ \scriptsize{$\pm 0.06$}} &
\makecell{\textbf{0.5656} \\ \scriptsize{$\pm 0.038$}} & \makecell{\textbf{0.5440} \\ \scriptsize{$\pm 0.039$}} &
\makecell{$0.1344$ \\ \scriptsize{$\pm 0.049$}} \\
UCN+LongCLIP & 20k & 1.8B & 57.7 & \makecell{$10.85$ \\ \scriptsize{$\pm 2.65$}} &\makecell{$0.1922$ \\ \scriptsize{$\pm 0.09$}} &  \makecell{$0.5704$ \\ \scriptsize{$\pm 0.05$}} & \makecell{\textbf{0.1449} \\ \scriptsize{$\pm 0.07$}} &
\makecell{$0.5508$ \\ \scriptsize{$\pm 0.041$}} & \makecell{$0.5735$ \\ \scriptsize{$\pm 0.046$}} &
\makecell{\textbf{0.1362} \\ \scriptsize{$\pm 0.055$}} \\
\bottomrule
\end{tabular}
\parbox{\linewidth}{\footnotesize FID for features extracted from an ImageNet-pretrained Inception V3. Smaller FID and bigger PSNR indicate higher fidelity to the original image distribution. Smaller diversity SSIM and FSIM indicate higher intra-prompt image generation diversity. Higher precision SSIM, MS-SSIM, FSIM and lower LPIPS indicate greater precision with ground-truth satellite images. Training setups use a low learning rate ($1e^{-5}$) while fine-tuning decoder with a lower learning rate ($2e^{-6}$). The best scores achieved through fine-tuning are in bold. Most values are provided as mean±standard deviation. CLIP, Contrastive Language-Image Pre-training (text encoder component); SD, Stable Diffusion (initialized with original weights); UCN, UrbanControlNet.}
\label{tab:main2}
\end{table}

Ideally, synthetic satellite images should closely match the distribution of the real images they are modelled after in terms of fidelity and precision, while at the same time capturing the full range of variability present in real-world imagery (diversity).

Image fidelity, precision and diversity were assessed using a set of complementary metrics, including Frechet Inception Distance (FID), Peak Signal-to-Noise Ratio (PSNR), Multi-Scale Structural Similarity Index (MS-SSIM), Learned Perceptual Image Patch Similarity (LPIPS), Structural Similarity Index (SSIM), and Feature Similarity Index (FSIM) (Table 1).  A lower FID indicates that the distribution of generated images is closer to that of the real ones in the feature space extracted by a pre-trained classification model (Inception v3), suggesting higher realism. PSNR quantifies the reconstruction quality based on pixel-wise differences, where a higher PSNR reflects better fidelity. LPIPS measures perceptual similarity using deep feature embeddings, with lower values indicating closer resemblance to real images. SSIM and FSIM focus on structural and feature-based similarities, respectively. Lower SSIM and FSIM in diversity imply greater diversity among generated samples, while higher SSIM and FSIM in precision imply greater precision with ground-truth satellite images.

These metrics should be jointly considered when comparing synthetic image quality. UrbanControlNet outperforms the baseline in all metrics (Table 1). Compared with the original Stable Diffusion model (FID of 86.5), all fine-tuned models, including runs with as few as 20k training steps (FID of 72.5, LPIPS of 0.5808), significantly improve fidelity and precision. ControlNet achieves a moderate performance (FID = 76.5) between the original stable diffusion (SD) and our proposed UrbanControlNet. Increasing the number of training steps to 60k led to better performance (e.g, lower FID of 50.3, higher precision SSIM of 0.1357, higher precision LPIPS of 0.5560, and lower diversity FSIM of 0.5520).

Switching the default general-domain CLIP text encoder with a domain-specific text encoder (e.g., RemoteCLIP \cite{remoteclip}) slightly improved performance when the text encoder was kept frozen. UrbanControlNet with RemoteCLIP shows superior FID (48.6), PSNR (11.23) and FSIM (0.5673), suggesting that text encoder adaptation can enhance fidelity without degrading visual quality. To address the 77-token limitation in the original CLIP, we further replaced the text encoder with a longer version (LongCLIP, 248 tokens).

After fine-tuning UrbanControlNet, we further fine-tune the SD decoder with fewer training steps (20k), and we find that performance improves further. In particular, UrbanControlNet with RemoteCLIP achieves the best FID (41.5),  PSNR (11.23), precision FSIM (0.5656) and precision LPIPS (0.5440) across all models. However, this comes with a slight trade-off in diversity (e.g., higher diversity FSIM  = 0.5696), as decoder fine-tuning tends to focus more on fidelity and precision over variation. It makes UrbanControlNet with RemoteCLIP a balanced choice for applications demanding both accuracy and diversity. Overall, full fine-tuning of both UrbanControlNet and the decoder provides a strong baseline.


\begin{table}[h]
\begin{center}
\caption{Prediction performance of a pre-trained CNN on $D_{test}$ and on synthetic SAIs generated using the models listed in the experiment column}
\begin{tabular}{@{}lcccc@{}}
\toprule
\textbf{Experiment} & \multicolumn{2}{c}{BCR}  & \multicolumn{2}{c}{BVD}    \\
 & MAE $\downarrow$ & $R^2\uparrow$ & MAE$\downarrow$ & $R^2\uparrow$ \\
\midrule
\textbf{UrbanControlNet} & 171.525 & 0.619 & 6.186 & 0.871 \\
\bottomrule
\end{tabular}
\label{tab:main1}
\end{center}
\end{table}

\section{Text-constraint-conditional counterfactual synthesis of urban satellite imagery }
To assess the alignment between the text prompts and the generated visual satellite concepts, a reader study was conducted for synthetic satellite images generated by the counterfactual prompt. If we increase the density value in the prompt, the synthetic satellite image will add more buildings (Figure S3). Similarly, the synthetic satellite image will reduce the number of buildings when we decrease the density value in the prompt. Interestingly, the building height will also increase in generated images, when the building volume density increases and the building coverage ratio achieves 100 \% in the prompt. This suggests that the trained our framework has effectively captured the distinctions among different density levels.

\begin{figure*}[!htbp]
\centering
\includegraphics[width=\textwidth]{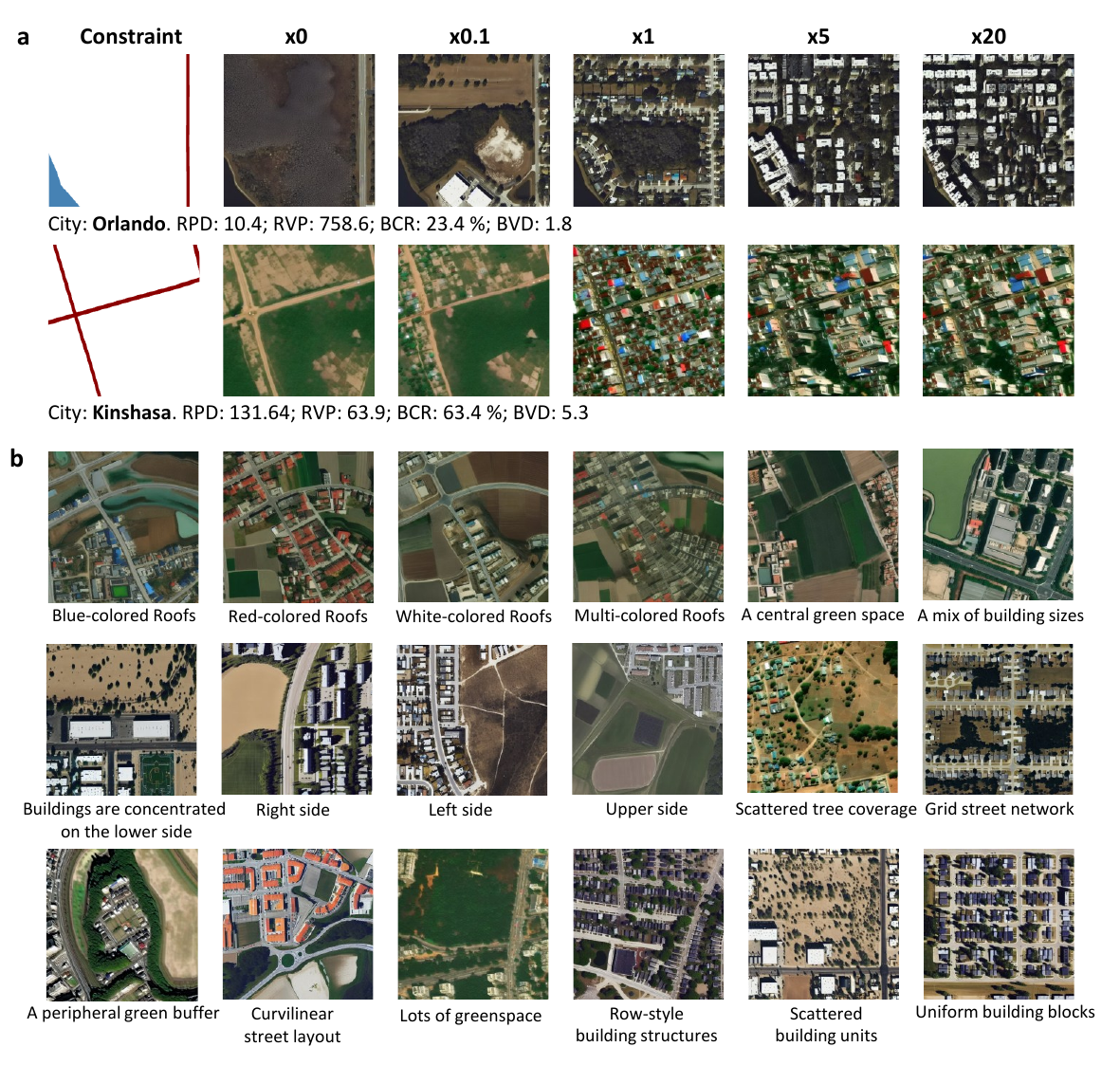}
\caption{\small{\textbf{Text-constraint-conditional counterfactual synthesis of urban satellite imagery .}  \textbf{a}.  Generated images respond to counterfactual density inputs. \textbf{b}. Generated images respond to free-form texts. }}
\label{text-multi}
\end{figure*}

\section{Style and content loss}
\label{styleloss}

Style loss emphasizes the stylistic differences between the generated image and the reference style image, often by comparing Gram matrices of deep features. In contrast, content loss measures the difference in content between the generated image and the original content image.

\subsection{Style Loss}

The style loss in our framework is defined using a deep convolutional neural network (VGG-19 pre-trained on ImageNet) \cite{DBLP:journals/corr/GatysEB15a}. For each generated image and each real reference image, we extract feature maps from multiple convolutional layers (conv1, ... ,conv5) and compute their Gram matrices, which capture correlations between feature channels and thus encode color and texture statistics.

The style loss is then calculated as the mean squared error (MSE) between the Gram matrices of the generated image and those of the real images, averaged across all considered layers. In our implementation, this corresponds to:

For each layer $\ell \in \mathcal{L}$ (with $\mathcal{L} = \{\text{conv1},\ldots,\text{conv5}\}$), let
$F_\ell(x) \in \mathbb{R}^{C_\ell \times H_\ell \times W_\ell}$ denote the feature map of image $x$.
We reshape it to $\tilde{F}_\ell(x) \in \mathbb{R}^{C_\ell \times N_\ell}$ with $N_\ell = H_\ell W_\ell$.
The (normalized) Gram matrix is:

\[
G_\ell(x) = \frac{1}{C_\ell N_\ell}\,\tilde{F}_\ell(x)\,\tilde{F}_\ell(x)^\top
\;\in\;\mathbb{R}^{C_\ell \times C_\ell}.
\]

The per-layer style loss between a generated image $g$ and a real image $r$ is:

\[
\ell_{\text{style}}^{(\ell)}(g,r)
= \frac{1}{4} \cdot \frac{1}{C_\ell^2}
\left\|\, G_\ell(g) - G_\ell(r) \,\right\|_F^2,
\]

where $\|\cdot\|_F$ denotes the Frobenius norm.

The total style loss averages over all common layers $\mathcal{L}$ and applies a scale factor of $10^4$:

\[
\mathcal{L}_{\text{style}}(g,r)
= 10^{4} \cdot \frac{1}{|\mathcal{L}|}
\sum_{\ell \in \mathcal{L}} \ell_{\text{style}}^{(\ell)}(g,r).
\]

Finally, the full experiment computes $\mathcal{L}_{\text{style}}(g,r)$ for all generated–real pairs
$(g,r) \in \mathcal{G} \times \mathcal{R}$.

\subsection{Content Loss}

The content loss in our framework is computed using a deep convolutional neural network (VGG-19 pre-trained on ImageNet). For each generated image, we extract its feature representation from the \texttt{conv\_4} layer of VGG-19, which captures mid-level semantic information including shapes and structures.

We compare this representation to the features of a set of sampled real images from the same city. After pooling both feature maps to a fixed spatial resolution, the contnet loss is defined as mean squared error between the generated features and the real features:

Let $f_{\text{conv\_4}}(\cdot)$
,  $\;\phi_{\ell}(x) \in \mathbb{R}^{C_\ell \times H_\ell \times W_\ell}$ denote the VGG19 feature map of image $x$ at layer $\ell = \text{conv4}$.
We apply adaptive average pooling to a fixed spatial resolution of $(32 \times 32)=1024$ and then flatten:

\[
f_{\text{conv\_4}}(x) = \operatorname{vec}\!\Big(\operatorname{AdaptiveAvgPool}_{(32,32)}\!\big(\phi_{\text{conv4}}(x)\big)\Big)
\;\in\;\mathbb{R}^{D}, \quad D = C_\ell  \times 1024.
\]

Given a set of sampled real images $\mathcal{R} = \{r_i\}_{i=1}^R$, the content loss for a generated image $g$ is defined as:

\[
\mathcal{L}_{\text{content}}(g;\mathcal{R})
= \frac{1}{R} \sum_{i=1}^{R}  \,\big\| f_{\text{conv\_4}}(g) - f_{\text{conv\_4}}(r_i) \big\|_2^2.
\]

\begin{figure*}[htb]
\centering
\includegraphics[width=\textwidth]{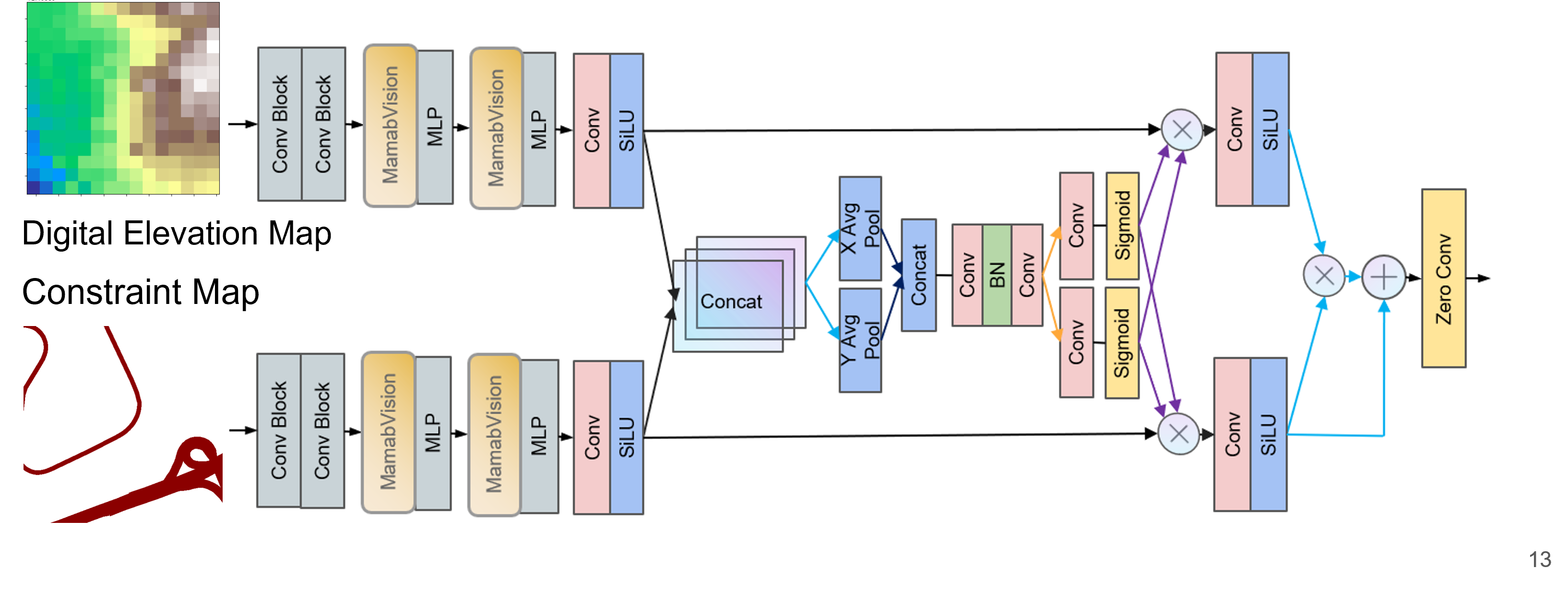}
\caption{\textbf{Multi-source pseudo-Siamese network in UrbanControlNet.}}
\label{vis-urban}
\end{figure*}

\begin{figure*}[htb]
\centering
\includegraphics[width=.5\textwidth]{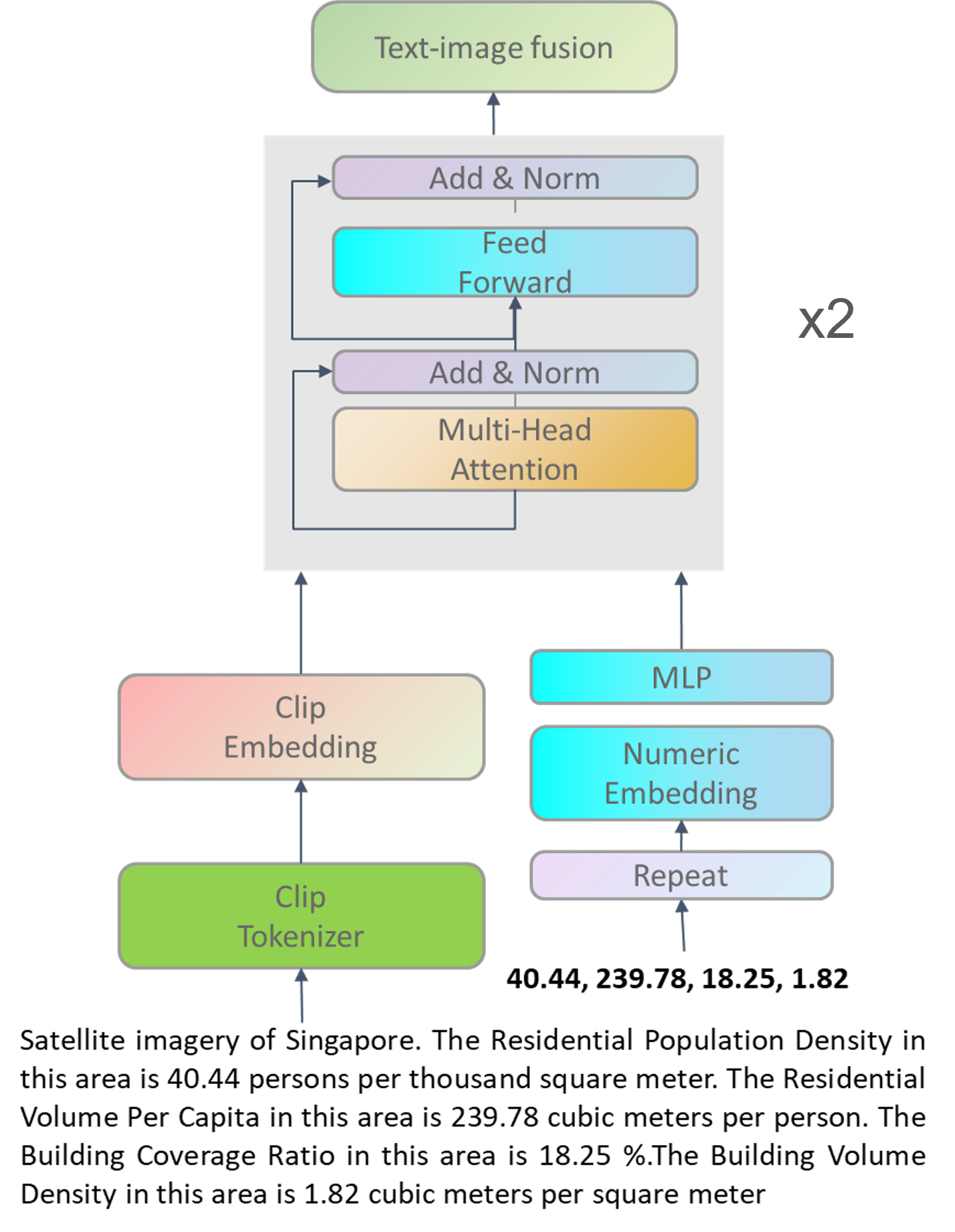}
\caption{\textbf{Numeral-aware network in UrbanControlNet.}}
\label{text-urban}
\end{figure*}

\begin{figure*}[!htbp]
\centering
\includegraphics[width=\textwidth]{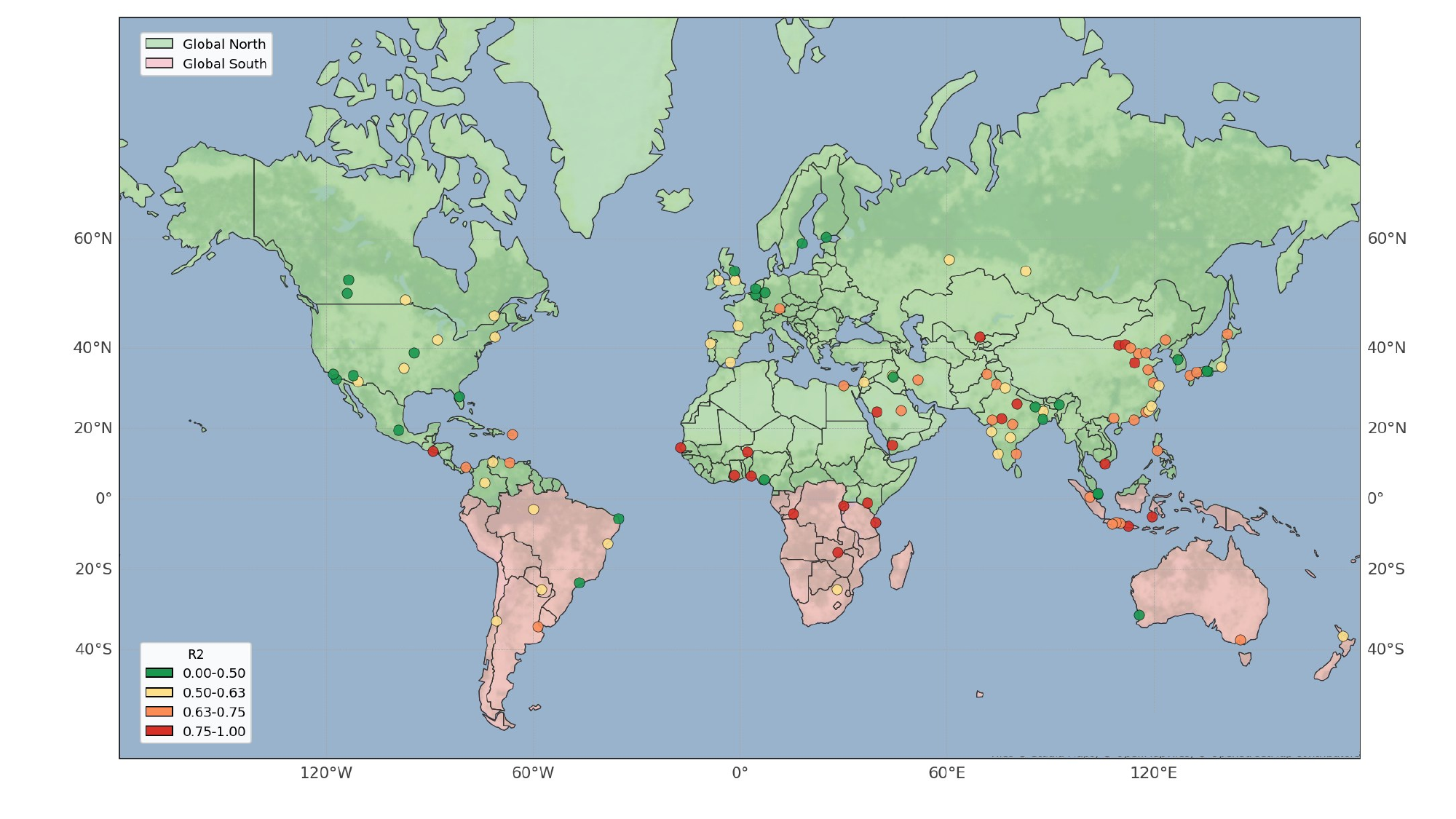}
\caption{\small{\textbf{ Model performance (R\textsuperscript{2}) distribution across more than 100 cities globally.}}}
\label{gs-r2}
\end{figure*}
This ensures that the generated images remain close in content to the real distribution while still allowing for stylistic differences.


\begin{table}[htbp]
\centering
\caption{Performance comparison of different models for global fossil fuel carbon emission prediction.}
\label{SI-CE}
\renewcommand{\arraystretch}{1.05}
\setlength{\tabcolsep}{2pt}
\begin{tabular}{lcccc}
\hline
\textbf{Model} & $R^2(\log)\uparrow$ & $R^2\uparrow$ & MAE$\downarrow$ & RMSE$\downarrow$ \\
\hline
ResNet & 0.830 & 0.610 & 22.04 & 60.82 \\
ResNet+TradAug & 0.837 & 0.630 & 21.74 & 57.72 \\
ResNet+AutoAug \cite{cubuk2020randaugment} & 0.842 & 0.626 & 21.45 & 58.53 \\
ResNet+SARandAug \cite{xiao2025sample} & 0.834 & 0.634 & 21.69 & 57.79 \\
\textbf{ResNet+GA} & \textbf{0.879} & \textbf{0.696} & \textbf{18.42} & \textbf{52.86} \\
\hline
Swin Transformer & 0.874 & 0.690 & 19.01 & 53.73 \\
Swin Transformer+TradAug & 0.872 & 0.697 & 18.99 & 52.23 \\
Swin Transformer+AutoAug \cite{cubuk2020randaugment} & 0.870 & 0.715 & 18.85 & 50.83 \\
Swin Transformer+SARandAug \cite{xiao2025sample} & 0.875 & 0.725 & 18.46 & 49.92 \\
\textbf{Swin Transformer+GA} & \textbf{0.903} & \textbf{0.764} & \textbf{15.86} & \textbf{46.19} \\
\hline
DINOv3 (SAT-493M) & 0.902 & 0.731 & 17.22 & 49.76 \\
DINOv3 (SAT-493M)+TradAug & 0.901 & 0.734 & 17.01 & 49.41 \\
DINOv3 (SAT-493M)+AutoAug \cite{cubuk2020randaugment} & 0.900 & 0.725 & 17.60 & 49.81 \\
DINOv3 (SAT-493M)+SARandAug \cite{xiao2025sample} & 0.904 & 0.742 & 17.03 & 48.55 \\
\textbf{DINOv3 (SAT-493M)+GA} & \textbf{0.926} & \textbf{0.776} & \textbf{14.65} & \textbf{45.22} \\
\hline
\end{tabular}
\end{table}

\section{Multi-source pseudo-Siamese Mamba network and numeral-aware network }
(1) Unlike the original ControlNet, the proposed  UrbanControlNet must handle double input maps (digital elevation map and road constraint map). To address it, we design a multi-source pseudo-Siamese network with Mamba networks and self-attention mechanisms in Fig.\ref{vis-urban}. First, the vision control network begins with separate pathways for the digital elevation map (DEM) and the constraint Map. Each pathway consists of convolutional blocks and MLPs that preprocess and extract initial spatial features. The state-of-the-art (SOTA) MambaVision block \cite{hatamizadeh2024mambavision} is designed to enhance spatial features based on the learned importance from these maps. We design a parallel hybrid architecture that combines the Selective State Model (SSM) from Mamba with Vision Transformers (ViT), specifically designed for vision-related tasks. This model enhances the ability to capture global context and long-range spatial dependencies by strategically integrating self-attention blocks in the later stages of the network. Next, features from both pathways are concatenated. To effectively learn their spatial correlation, this study employs a coordinate attention block with X,Y average pool layers to learn their aggregated features. We use the Sigmod function after the
 aggregated feature to construct a learnable special mask (the value of this mask belongs to (0,1).  This mask aims to predict the spatial score
 of the feature maps so that more effective features can be activated. A self-attention mechanism with a Gate network is designed to enhance the joint features between digital elevation features and road constraint features. Then, their individual and aggregated features are multiplied by element-wise (Hadamard) product and addition in Eq (1):
\begin{equation}
\bm{h}_{{\rm agg}}=\alpha \cdot \bm{h}_{{\rm dem}} \odot \bm{h}_{{\rm constraint}}+\beta \cdot \bm{h}_{{\rm constraint}}.
\end{equation}
Last, the final output is processed through a "zero convolution" layer to ensure that the network's output matches the stable diffusion. The "zero convolution" is 1×1 convolution with both weight and bias initialized as zeros. Before training, all zero convolutions output zeros, and UrbanControlNet will not cause any distortion. (2) Conventional methods primarily condition image generation on captions or images but struggle with numerical constraints, such as density metrics. To address this,  we introduce a numeral-aware network to enhance the text embedding in Fig.\ref{text-urban}, which effectively integrates numerical data as conditioning information. The network begins with the pre-trained Clip tokenizer and embedding, which encode textual data. To enhance the numeric information, we add numeric embeddings and MLP to encode the numeric data and learn the numeric features. Then, we utilize the multi-layer cross-attention network to extract the correlation between text features and numeric features. Last, the aggregated features will be fused with the above multi-source pseudo-Siamese network for future text-image fusion.

\FloatBarrier

\end{document}